\PassOptionsToPackage{table}{xcolor}
\documentclass[]{template/fairmeta}
\usepackage{amsmath,amssymb}
\usepackage{graphicx}
\usepackage[numbers]{natbib}
\usepackage{booktabs}
\usepackage{bm}
\usepackage{xspace}
\usepackage{tabularx}
\usepackage{geometry}
\usepackage{multirow}
\usepackage{enumitem}
\usepackage{iftex}
\usepackage{titlesec}
\usepackage{subcaption}
\usepackage{algorithm}
\usepackage{algpseudocode}

\usepackage[
  hidelinks,
  breaklinks,
  colorlinks=true,
  linkcolor=blue,
  citecolor={blue!80!black},
  urlcolor=cyan
]{hyperref}

\ifPDFTeX
  \PackageError{paper}{This document contains CJK text and must be built with
  XeLaTeX}{Run `latexmk -pdfxe paper.tex` or `make`.}
\else
  \usepackage{fontspec}
  \usepackage{xeCJK}
  \IfFontExistsTF{Arial Unicode MS}{
    \setCJKmainfont{Arial Unicode MS}
    \setCJKmonofont{Arial Unicode MS}
  }{
  }
  \newfontfamily\calmregular{Outfit-Regular.ttf}[
    BoldFont=Outfit-Regular.ttf
  ]
  \newcommand{\calmfont}[1]{{{\calmregular #1}}}
\fi

\newcommand{\modelname}{EmbodiedGen V2\xspace}
\newcommand{\eg}{e.g.\xspace}

\titleformat{\section}
  {\normalfont\Large\bfseries\calmregular}{\thesection}{1em}{}
\titleformat{\subsection}
  {\normalfont\large\bfseries\calmregular}{\thesubsection}{1em}{}
\titleformat{\subsubsection}
  {\normalfont\normalsize\bfseries\calmregular}{\thesubsubsection}{1em}{}
\titleformat{\paragraph}[runin]
  {\normalfont\normalsize\bfseries\calmregular}{\theparagraph}{1em}{}

\title{\calmfont{\modelname: An Agentic, Simulation-Ready \\
3D World Engine for Embodied AI}}

\author{\calmfont{Xinjie Wang$^{1}$, Liu Liu$^{1}$, Taojun Ding$^{1}$, Andrew Choi$^{1}$, Chaodong Huang$^{1}$, Mengao Zhao$^{1}$, Ziang Li$^{1}$, Jackson Jiang$^{2}$, Chunlei Yu$^{2}$, Shengxiang Liu$^{2}$, Wei Xu$^{1}$, Zhizhong Su$^{1}$}}

\affiliation[1]{\calmfont{Horizon Robotics}}
\affiliation[2]{\calmfont{WuwenAI}}
\code{\url{github.com/HorizonRobotics/EmbodiedGen}}
\webpage{\url{horizonrobotics.github.io/EmbodiedGen}}

\begin{document}

\abstract{
  We present \modelname, a generative 3D world engine for building executable policy-ready environments for embodied intelligence.
  Sim-ready 3D asset generation has advanced rapidly, yet assembling such assets into policy-ready task environments remains largely manual, limiting scalable closed-loop learning.
  \modelname addresses this gap through a unified sim-ready representation that connects cross-simulator assets, interaction affordances, task-driven worlds, large-scale multi-room scenes, and stateful Vibe Coding into a generative, editable, and reusable simulation pipeline.
  The generated environments support manipulation, navigation, mobile manipulation, cross-simulator deployment, and embodied policy training.
  In evaluation, the asset pipeline achieves 96.5\% human acceptance and 98.6\% collision success, and 83.3\% of task-driven worlds are directly usable for downstream simulation without manual modification.
  Online reinforcement learning with generated environments further improves simulation success from 9.7\% to 79.8\%, and transfers to real robots with task success increasing from 21.7\% to 75.0\%.
  These results establish \modelname as scalable simulation infrastructure for training, evaluating, and deploying embodied policies.
}

\vspace*{-2.5cm}
\maketitle

\begingroup
\captionsetup{font=small,skip=2pt,hypcap=false}
\vspace{-0.7em}
\noindent\makebox[\linewidth][c]{%
  \includegraphics[width=1.0\linewidth]{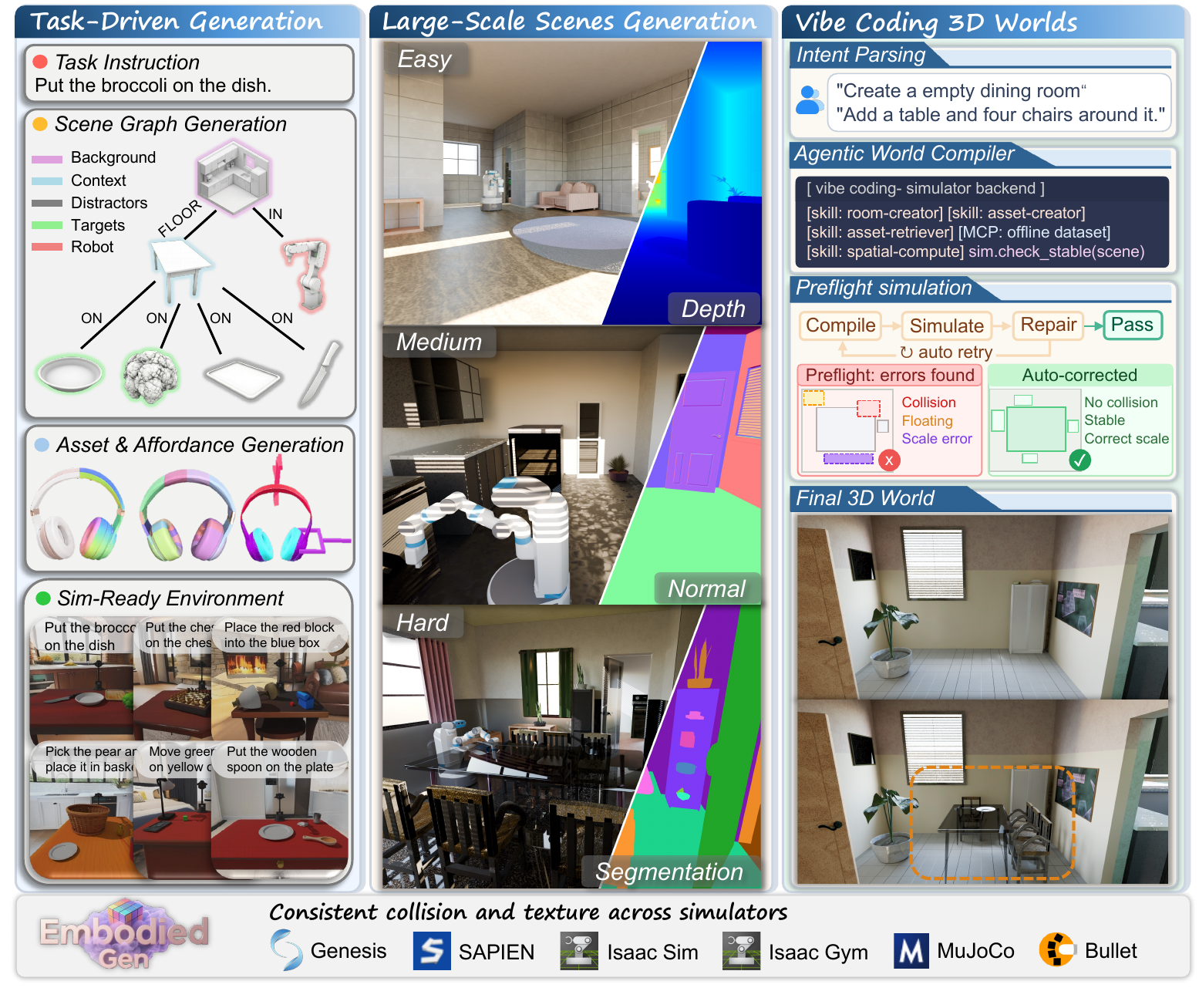}%
}
\vspace{-2.3em}
\captionof{figure}{
  Overview of \modelname.
  \textbf{Left:} natural-language task to sim-ready scene via Scene Graph
  and affordance-annotated assets.
  \textbf{Middle:} large-scale multi-room generation at different
  controllable complexity tiers.
  \textbf{Right:} Vibe Coding 3D world editing.
  All outputs deploy consistently across mainstream simulators.
}
\label{fig:overview}
\endgroup

\clearpage

\section{Introduction}
\label{introduction}

Generative 3D models have made rapid progress in producing visually plausible objects and scenes, but embodied policy learning requires more than visual 3D content.
Robots and embodied agents need executable task environments: objects must carry physical and interaction properties, layouts must satisfy task and navigation constraints, and the resulting worlds must be portable across simulators, editable, and usable for closed-loop training and evaluation.
We refer to environments that satisfy these requirements and can be used in physics simulation without manual adaptation as \emph{sim-ready}.
Although recent pipelines have made important progress on simulation-ready assets and generative 3D worlds, generating scalable sim-ready environments for embodied tasks remains challenging.
The key difficulty is to preserve geometry, physics, affordances, task semantics, and simulator interfaces within a unified world representation rather than treating them as separate post-processing steps.

Building on EmbodiedGen V1~\cite{wang2025embodiedgengenerative3dworld}, we introduce \modelname, which advances from a generative 3D content toolkit to a sim-ready 3D world engine for embodied task-environment generation, policy learning, and evaluation.
The system is organized around complete executable environments rather than isolated generated artifacts.
For large-scale world generation, \modelname generates structured multi-room and whole-house scenes with explicit room topology, traversable openings, and individually addressable furniture, overcoming the limited camera translation of V1's panorama-back-projected single-mesh backgrounds and enabling long-horizon navigation and mobile manipulation.
For asset and interaction generation, the asset pipeline becomes pluggable across TRELLIS~\cite{xiang2024structured}, SAM3D~\cite{sam3dteam2025sam3d3dfyimages}, and Hunyuan3D~\cite{hunyuan3d2025hunyuan3d}, accepts partially occluded images through in-place scene completion~\cite{yin20263dfixer}, augments objects with part-level affordances and physically validated grasps, and extends deployment to deformable bodies.
For deployment and editing, \modelname standardizes export through URDF, simulator XML formats including MJCF, and USD for mainstream simulators, while Vibe Coding provides stateful natural-language editing over a persistent, physics-validated world state.
Finally, the system is evaluated beyond static generation quality: generated environments support online reinforcement learning of VLA policies and sim-to-real policy transfer~\cite{choi2026scaling,choi2026rankq}.

Together, these capabilities connect generative 3D world generation to the main workflows of embodied policy development.
By preserving simulator portability, interaction semantics, task-conditioned layouts, and editable world state in one representation, \modelname supports cross-simulator reuse, policy training, controlled environment variation, policy debugging, and closed-loop evaluation.

Our main contributions are summarized as follows:

\begin{itemize}[leftmargin=*]
  \item \textbf{\calmfont{Unified sim-ready representation.}}
    We introduce a shared representation that couples metric geometry, physical validity, interaction semantics, and cross-simulator portability, providing a common interface across the full generation and editing pipeline.

  \item \textbf{\calmfont{Model-agnostic simulation asset generation.}}
    We build a modular text/image-to-3D asset pipeline that plugs TRELLIS~\cite{xiang2024structured}, SAM3D~\cite{sam3dteam2025sam3d3dfyimages}, and Hunyuan3D~\cite{hunyuan3d2025hunyuan3d} into a unified post-processing stack, converting raw 3D outputs into deployable simulation assets via quality checking, mesh repair, convex decomposition, texture baking, physical property recovery, and cross-simulator export.

  \item \textbf{\calmfont{Affordance-aware interaction semantics.}}
    We introduce an affordance autolabeling pipeline that augments generated assets with semantic interaction attributes, turning visual and collision-aware objects into actionable, verifiable simulation entities. We further release a 4K+ asset collection with cross-simulator formats and affordance annotations.

  \item \textbf{\calmfont{Task-driven world and scene generation.}}
    We parse open-ended natural-language tasks into Scene Graphs and generate executable interactive worlds through spatial constraints and physical stability solving; we further extend the same representation to large-scale scenes with multi-room topology, navigable space, and instance-level editability.

  \item \textbf{\calmfont{Stateful Vibe Coding for 3D world editing.}}
    We expose the shared world representation through a stateful agent--skill harness, turning generation, composition, editing, and export into composable skills so that natural-language instructions become bounded, physics-validated edits over a persistent, deployable world state.

  \item \textbf{\calmfont{Closed-loop policy validation.}}
    We validate the generated environments beyond static asset and scene quality: downstream VLA/RL studies~\cite{choi2026scaling} show that online fine-tuning purely with \modelname-generated environments improves simulation success from 9.7\% to 79.8\% and real-robot task success from 21.7\% to 75.0\%.
\end{itemize}

\section{Methodology}
\label{methods}

\subsection{Preliminaries}
\label{sec:preliminaries}

We target \emph{simulation-ready 3D world generation}: producing worlds that are not only visually plausible, but also directly executable by embodied agents in physics simulation.
Throughout this paper, we use \emph{sim-ready} to denote an output contract that couples four requirements: metric geometry, simulation-compatible physical assets, task-level semantics and affordances, and standardized simulator interfaces.

\modelname instantiates this contract with a two-level representation.
At the object level, each sim-ready asset bundles textured visual geometry, collision geometry, physical parameters, and affordance annotations.
At the scene level, a typed Scene Graph specifies entities, task roles---background, contexts, manipulated objects, distractors, and the robot---and their spatial and interaction relations, which are then grounded into physically stable 6-DoF poses in a target simulator.
The following modules build this representation progressively: sim-ready asset generation creates deployable objects, affordance autolabeling enriches them with interaction semantics, task-driven interactive worlds generation and large-scale scenes generation compose executable worlds, and Vibe Coding exposes the same representation as a stateful natural-language editing interface.

\subsection{Sim-Ready 3D Asset Generation}
\label{sec:sim_ready_asset}

\begin{figure}[t]
  \centering
  \includegraphics[width=0.95\linewidth]{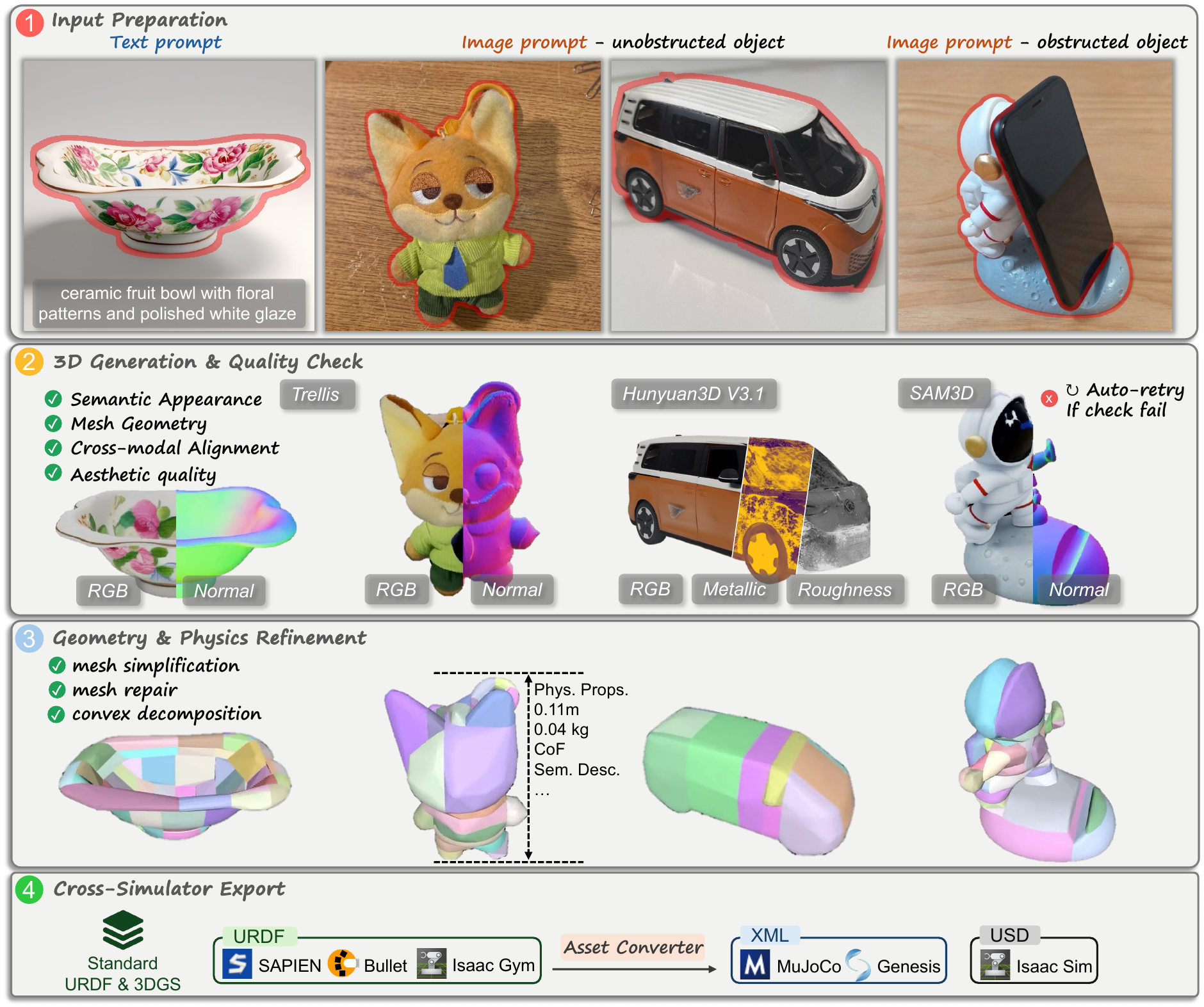}
  \caption{
    The sim-ready 3D asset generation pipeline. From text or image inputs, the
    system produces simulation-ready assets through input preparation, 3D
    generation, geometry refinement \& texture baking, VLM-driven physical
    property recovery, and cross-simulator export.
  }
  \label{fig:sim_ready_asset_gen}
\end{figure}

\textbf{Problem definition.}
Existing image-to-3D and text-to-3D generative models~\cite{xiang2024structured,zhao2025hunyuan3d20,li2025step1x3d,%
lai2025hunyuan3d25,xiang2025trellis2} can produce visually plausible 3D objects, yet their outputs typically possess only \emph{visualization-level 3D usability} rather than \emph{simulation-level asset usability}: meshes reside in a normalized coordinate space, lack real-world scale, mass, and friction properties, may contain non-manifold faces or open surfaces, and carry neither collision representations nor standardized simulation interfaces.
We define \emph{sim-ready asset generation} as the task of automatically producing, from open-ended text or image conditions, object assets that jointly contain simulation-compatible geometry, explicit textures, structured physical metadata, and standardized simulation representations consumable by different physics engines.

\textbf{Pipeline overview.}
As shown in Fig.~\ref{fig:sim_ready_asset_gen}, we propose a unified sim-ready asset generation pipeline that maps three types of input---text prompts, unoccluded object images, and partially occluded object images---into standardized simulation assets through five stages:
\textbf{(i)~Input preparation}: for text input, a pluggable text-to-image model (SD3.5~\cite{esser2024scaling} or Kolors~\cite{kolors}) generates candidate object images; for image input (including occluded scenes), a foreground segmentation model (Rembg~\cite{Gatis_rembg_2025}, SAM~\cite{kirillov2023segany}, or RMBG~\cite{briaai_rmbg_1_4}) extracts the target object.
\textbf{(ii)~3D generation}: the foreground image is fed into a pluggable image-to-3D model (TRELLIS~\cite{xiang2024structured}, SAM3D~\cite{sam3dteam2025sam3d3dfyimages}, or Hunyuan3D~\cite{hunyuan3d2025hunyuan3d}), yielding both a 3D Gaussian~\cite{kerbl20233dgaussiansplattingrealtime} and a mesh as intermediate representations.
\textbf{(iii)~Geometry refinement \& texture baking}: the mesh undergoes topological repair and simplification, while multi-view back-projection bakes the Gaussian appearance into an explicit texture map.
\textbf{(iv)~Physical property recovery}: a vision-language model (VLM) infers real-world scale, mass, and friction coefficients from multi-view renderings, which are then used for metric rescaling.
\textbf{(v)~Simulation asset packaging \& cross-format export}: geometry, appearance, and physical information are assembled into a unified intermediate representation and automatically converted to different simulator formats.
Crucially, we do not treat simulation compatibility as a separate export step appended after generation; instead, simulation-oriented quality constraints are explicitly enforced at multiple stages---candidate screening, 3D generation, and physical recovery---forming a closed-loop \emph{generate--verify--retry} pipeline.

\textbf{Hierarchical quality gating.}
Rather than relying on a single forward pass to produce the final result, we embed quality gates at multiple pipeline stages.
At the input stage, candidate images must pass a foreground quality check (a VLM validates semantic correctness and geometric completeness of the segmentation); for the text-driven path, the system additionally verifies that the generated image is semantically consistent with the original text intent.
Only qualified candidates proceed to 3D generation.
At the 3D generation stage, the system evaluates geometric integrity from multi-view renderings, rejecting truncated geometry, duplicate bodies, and extraneous attached elements; failures trigger automatic retries with different random seeds.
At the pipeline's end, an aesthetic scoring model~\cite{schuhmann_aesthetic_2025} quantitatively rates texture and geometry quality, filtering samples below a predefined threshold.
We write all quality-check results as structured tags into the final asset file, making each asset's quality status queryable and filterable in downstream large-scale usage.

\textbf{Geometry processing \& physical property recovery.}
Raw meshes from 3D generative models typically contain non-manifold faces and open regions that are incompatible with collision detection and physics simulation.
After topological repair and simplification, we apply the CoACD algorithm~\cite{wei2022coacd} to compute an approximate convex decomposition, producing a set of compact convex collision bodies; if decomposition fails, the pipeline falls back to the original mesh.
In Fig.~\ref{fig:sim_ready_asset_gen}, different colors indicate the decomposed mesh parts produced by this convex decomposition.
Since Gaussian representations are unsuitable as texture carriers for simulation assets, we bake the Gaussian model's multi-view appearance into a high-resolution explicit texture map via automatic UV unwrapping and differentiable rasterization.
For physical properties, a VLM infers semantic real-world scale, mass, and friction ranges from multi-view renderings and the object category.
We use the estimated scale to calibrate the visual mesh, collision mesh, and Gaussian representation consistently, then store the recovered mass and friction in the asset's inertial and contact metadata for downstream sampling or calibration.

\textbf{Cross-simulator asset export.}
We adopt URDF as the unified intermediate representation, as it natively supports structured packaging of visual meshes, collision meshes, inertial parameters, and auxiliary metadata, making it a suitable canonical representation across simulation back-ends.
A format converter automatically transforms the URDF (SAPIEN~\cite{Xiang_2020_SAPIEN}, Bullet~\cite{coumans2021}, Isaac Gym~\cite{makoviychuk2021isaac}) into XML (MuJoCo~\cite{todorov2012mujoco}, Genesis~\cite{Genesis}) and USD (Isaac Sim~\cite{NVIDIA_Isaac_Sim}), correctly handling visual/collision geometry separation, local coordinate transforms, material mapping, and physics property injection.
This design decouples \emph{object generation} from \emph{simulator adaptation}, enabling the same generated asset to be instantiated across different embodied simulation platforms with consistent physical behavior.
We provide online usage examples in RoboVerse~\cite{geng2025roboverse}.\footnote{\url{https://roboverse.wiki/metasim/get_started/quick_start/14_real_asset}}

\begin{figure}[t]
  \centering
  \includegraphics[width=1.0\linewidth]{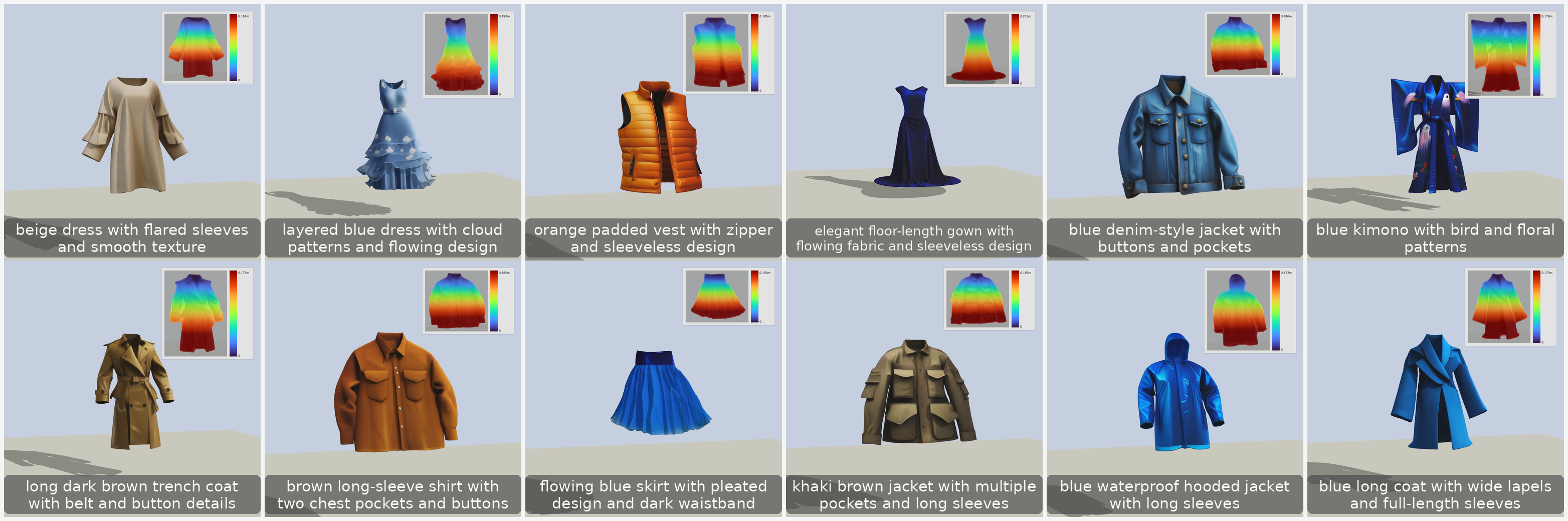}
  \caption{
    Twelve text-conditioned garments deployed as deformable meshes in Genesis~\cite{Genesis}.
    Inset heatmaps show per-vertex displacement under cloth dynamics, confirming
    that generated geometry supports soft-body simulation without manual preparation.
  }
  \label{fig:deformable_sim}
\end{figure}

\textbf{Deformable-body simulation.}
Beyond rigid-body physics, the same generation and export path extends naturally to soft-body simulation.
Fig.~\ref{fig:deformable_sim} shows twelve garments generated from text prompts and exported to Genesis~\cite{Genesis} as deformable meshes; per-vertex displacement heatmaps confirm that the generated surface geometry carries sufficient fidelity for cloth and soft-body dynamics without any manual mesh preparation.

\textbf{Single-image in-place completion.}
To handle partially occluded inputs, we use 3D-Fixer~\cite{yin20263dfixer}, which recovers complete object geometry in place by treating the fragmented visible point cloud as a spatial anchor.
It conditions a frozen TRELLIS backbone~\cite{xiang2024structured} through occlusion-robust feature alignment and coarse-to-fine completion, avoiding explicit pose optimization.

\subsection{Affordance Autolabeling Pipeline}
\label{sec:affordance_pipeline}

\textbf{Problem definition.}
Embodied manipulation requires semantics not only for object categories or instances, but also for interaction-relevant parts.
A manipulation policy must infer \emph{where} to make contact, \emph{what} function the contacted region supports, and \emph{how} the contact can be executed under geometric and physical constraints~\cite{deng20213daff,mo2021where2act}.
We define \emph{affordance autolabeling} as converting a sim-ready 3D asset into a structured part-level interaction representation: each mesh face receives a part identifier, and each part is annotated with its semantic name, graspability, task-relevant grasp scenarios, functional labels, appearance semantics, and simulation-validated candidate grasps.
This representation connects language-level intent to localized 3D contact regions and executable robot actions.

\begin{figure}[H]
  \centering
  \includegraphics[width=1.0\linewidth]{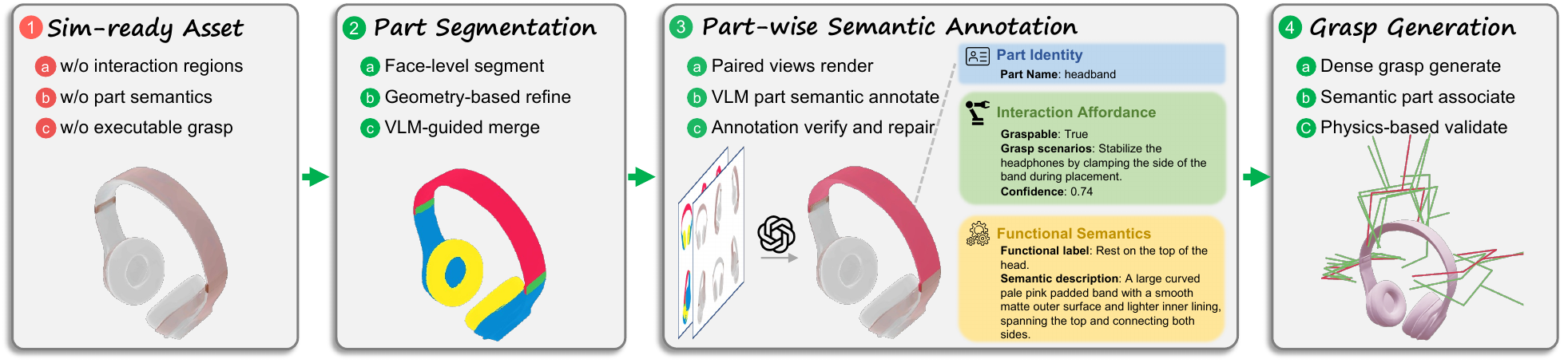}
  \caption{
    The affordance autolabeling pipeline. From sim-ready assets, the
    system produces structured part-level interaction representations through functional part segmentation, part-wise semantic annotation, and grasp generation \& physical validation.
  }
  \label{fig:affordance_ann_pipeline}
\end{figure}

\textbf{Pipeline overview.}
Starting from the sim-ready assets in Sec.~\ref{sec:sim_ready_asset}, our affordance autolabeling pipeline augments each generated object with a structured part-level schema for embodied manipulation.
As illustrated in Fig.~\ref{fig:affordance_ann_pipeline}, the pipeline proceeds through three stages:
\textbf{(i)~Functional part segmentation}: P3-SAM~\cite{ma2025p3sam} decomposes the mesh into functionally meaningful part regions, which serve as geometric carriers for affordance annotation;
\textbf{(ii)~Part-wise semantic annotation}: conditioned on the part masks, a VLM (GPT-5.4~\cite{openai2026gpt5dot4}) interprets aligned RGB and mask renderings to infer each region's semantic name, functional role, graspability, task-conditioned grasp scenarios, and appearance description; and
\textbf{(iii)~Grasp generation and physical validation}: GraspGen~\cite{murali2025graspgen} proposes 6-DoF grasp candidates, which are associated with contacted mesh parts and filtered through physics-based execution tests.
The resulting face-level part segmentation mesh and part-wise affordance annotations provide a simulator-grounded action interface, specifying which part an instruction refers to, what interaction function the part supports, and how the robot can execute the intended contact through simulation-filtered grasp groups.

\textbf{Functional part segmentation.}
For each sim-ready asset, P3-SAM~\cite{ma2025p3sam} decomposes the mesh into potentially functional part regions.
It samples a point cloud from the mesh, infers the object's part structure in normalized 3D space, and projects the predicted point-cloud masks back onto the original mesh faces to obtain a face-level part segmentation map.
The predicted part identifiers are remapped to a fixed color palette, giving the subsequent VLM stages directly accessible color names.
Raw P3-SAM predictions can still contain boundary errors and overly fine-grained partitions, so we add geometry-consistent post-processing and VLM-guided part merging.
The geometry-based post-processing corrects local labels by merging smoothly connected face components and relabeling small surrounded fragments, targeting projection noise without collapsing genuine sharp part boundaries.
For semantic over-segmentation, a VLM-based checker takes the object category, all part color names, a $2{\times}3$ multi-view grid of RGB renderings, and the aligned part-mask grid as input.
When the checker identifies that the same functional part has been split into multiple independent part regions, the pipeline automatically merges the corresponding regions and iterates this check-and-merge process until no further merges are needed.
The accepted output is a face-level segmentation map that is both geometrically continuous and semantically aligned with functional object structure.

\textbf{Part-wise semantic annotation.}
We further extract interaction-oriented semantics from the geometric part regions.
We reuse the segmentation-checker inputs: object category, part color names, and aligned RGB and part-mask renderings.
The RGB views provide appearance, material, and structure cues, while the masks preserve region identities across viewpoints, enabling the VLM to associate visual cues with the same 3D part.

These inputs are provided to a VLM (GPT-5.4~\cite{openai2026gpt5dot4}) to infer structured attributes for each physically meaningful part, including \emph{part name}, \emph{graspability}, task-conditioned \emph{grasp scenarios}, \emph{functional labels}, and a fine-grained \emph{semantic description}.
Graspability and grasp scenarios specify whether the region is suitable as a robotic contact target; functional labels characterize the part's role; and the semantic description records appearance-level cues such as color, material, texture, shape, and relative location.
The VLM response uses mask color names to associate each part-level annotation with its corresponding segmented region.
We further apply a VLM-based checker to judge and revise the response, producing the final affordance annotation.
The verified part-wise semantics form queryable part-level priors for downstream task planning.

\textbf{Grasp generation and physical validation.}
GraspGen~\cite{murali2025graspgen} generates confidence-scored 6-DoF grasp candidates, which we map to the contacted semantic parts and rank by confidence.
We validate these candidates in SAPIEN~\cite{Xiang_2020_SAPIEN} through simulated closing, lifting, perturbation, and lowering, retaining only grasps that remain stable relative to the gripper.
The resulting annotations pair part-level interaction semantics with physically executable grasp proposals for downstream manipulation.

\subsection{Task-Driven Interactive Worlds Generation}
\label{sec:layout_generation}

\textbf{Problem definition.}
We formulate task-driven interactive worlds generation as the task of mapping a natural-language task description (\eg, ``Place the fruit onto the plate on the table'') to a fully instantiated 3D world that is directly usable for simulation and satisfies execution constraints.
The output comprises two complementary representations: (i)~a \emph{Scene Graph}---a rooted multiway tree whose nodes correspond to 3D assets and whose edges encode spatial parent--child relationships---and (ii)~a composed interactive 3D world with real-scale geometry, physical properties, and 6-DoF poses for every object, directly loadable into physics simulators.
Unlike prior scene generation methods that take a room category or object list as input~\cite{yang2024holodeck,feng2023layoutgpt}, our formulation is \emph{task-driven}: the system autonomously reasons about which objects are needed, how they relate spatially, and where a robot should be placed to execute the described manipulation task.
This factorization is analogous to green-screen production in filmmaking: instead of jointly generating every world detail, we model an interactive environment as a background plus the minimal set of task-relevant interactive assets.
The abstraction preserves the semantic and physical constraints required by the task while substantially reducing world synthesis complexity and rendering cost.

\begin{figure}[H]
  \centering
  \includegraphics[width=1.0\linewidth]{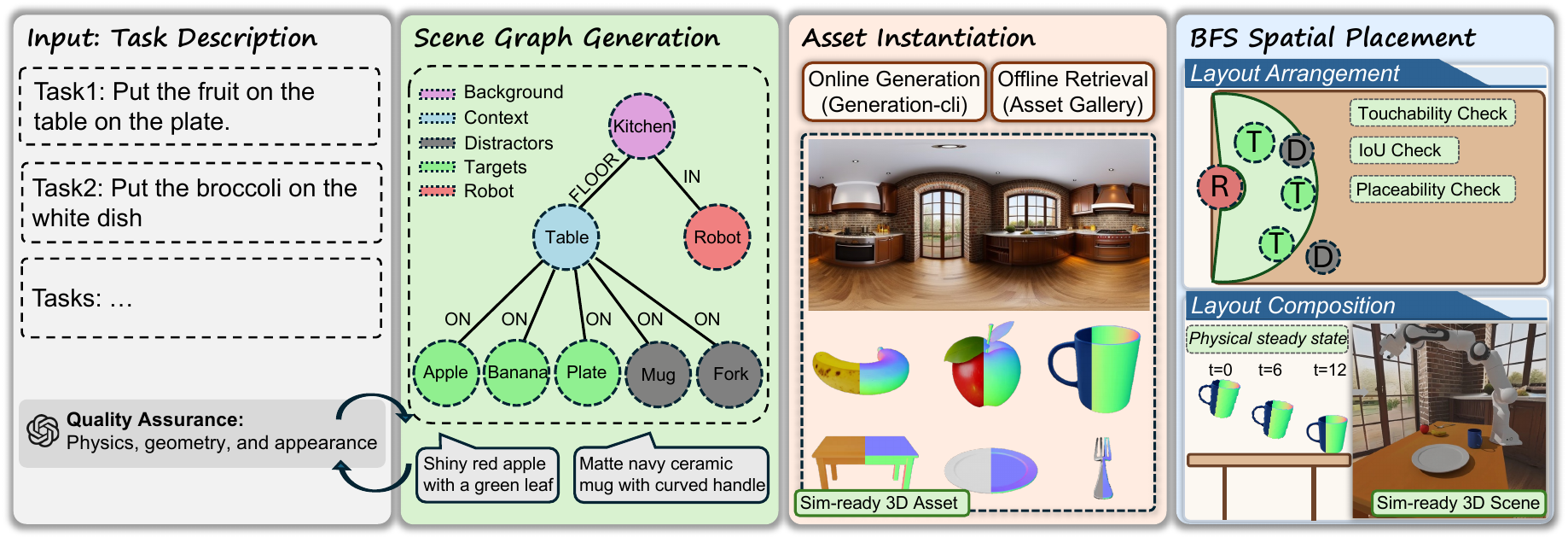}
  \caption{
    Task-driven interactive worlds generation pipeline: scene graph generation
    from a natural-language task, sim-ready asset instantiation, and BFS-based
    spatial placement with physics settling.
  }
  \label{fig:layout_pipeline}
\end{figure}

\textbf{Pipeline overview.}
As shown in Fig.~\ref{fig:layout_pipeline}, given a natural-language task description, the pipeline proceeds in three stages:
\textbf{(1)}~\emph{scene graph generation} parses the task into semantic roles, organizes them into a Scene Graph with explicit spatial relations, and generates per-object visual descriptions;
\textbf{(2)}~\emph{asset generation} instantiates each node as a sim-ready 3D asset; and
\textbf{(3)}~\emph{spatial placement} computes collision-free, physically stable poses via BFS traversal and physics settling.

\textbf{Scene Graph Generation.}
\emph{Scene decomposition.}
Given a task description, we prompt an LLM to decompose it into five semantic categories:
\emph{ROBOT}, denoting the robot type;
\emph{BACKGROUND}, the indoor environment;
\emph{CONTEXT}, the primary piece of furniture that anchors the interaction;
\emph{TARGETS}, the objects the robot must act upon; and
\emph{DISTRACTORS}, plausible scene props unrelated to the task.
The decomposition enforces semantic consistency: the context object must plausibly belong to the background.
For example, a kitchen counter is compatible with a kitchen but not a bedroom.
The decomposition also restricts outputs to rigid bodies suitable for physics simulation.

\emph{Hierarchy generation \& asset description.}
A second LLM query organizes the decomposed elements into a shallow rooted Scene Graph: the background is the root, the context and robot are its children, and manipulated and distractor objects attach to the context.
Edges encode the \texttt{ON}, \texttt{INSIDE}, \texttt{FLOOR}, and \texttt{IN} relations, while the single-parent structure reduces placement ambiguity.
Each asset node also receives a visual description that conditions the subsequent generation stage.

\textbf{Asset acquisition.}
To balance diversity and efficiency, both assets and backgrounds support two sourcing modes: online generation and offline database retrieval.
Each asset node in the Scene Graph can be instantiated on demand using the text-to-3D pipeline described in Sec.~\ref{sec:sim_ready_asset}, or retrieved from a pre-built asset library when a suitable instance already exists.
All instantiated assets carry real-world scale, mass, friction, collision geometry, and URDF packaging.
Background scenes likewise support both online generation and offline retrieval: we generate simple backgrounds with the background generation method of EmbodiedGen~\cite{wang2025embodiedgengenerative3dworld}, while the large-scale scenes generation module (Sec.~\ref{sec:large_scale_scene_generation}) produces complex backgrounds.

\begin{figure}[t]
  \centering
  \includegraphics[width=0.95\linewidth]{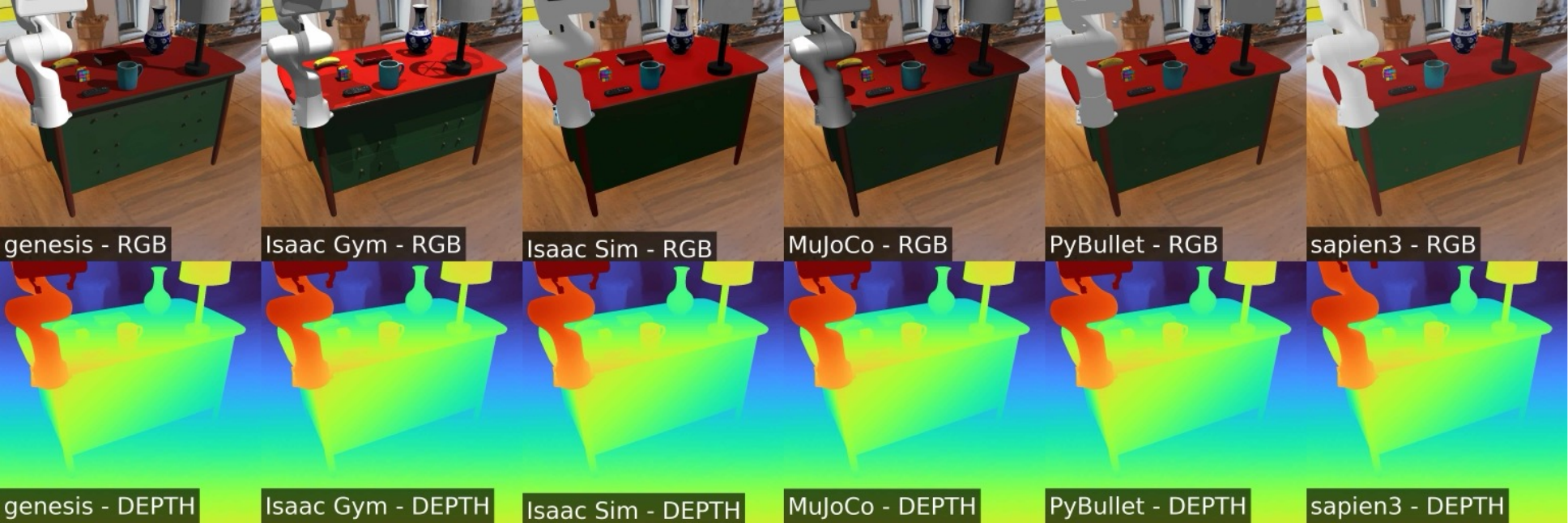}
  \caption{
    The same task-driven interactive world layout instantiated across six physics
    simulators
    (Genesis, Isaac Gym, Isaac Sim, MuJoCo, PyBullet, and SAPIEN3), shown in
    RGB (top) and depth (bottom). The standardized layout description produced
    by our pipeline requires no manual adaptation to be loaded and executed in
    each back-end, demonstrating cross-simulator portability.
  }
  \label{fig:cross_sim}
\end{figure}

\textbf{BFS spatial placement.}
We instantiate the Scene Graph in breadth-first order so that each parent is placed before its children, sorting siblings by footprint to reserve support space for larger objects.
For each spatial relation, a relation-specific sampler proposes a child position $\mathbf{p}_c$ on or within the parent geometry subject to support and collision constraints:
\begin{equation}
\label{eq:placement}
\mathbf{p}_c \in \mathcal{H}_p, \qquad
\mathrm{Support}\!\left(\mathcal{B}_c(\mathbf{p}_c), \mathcal{H}_p\right)=1, \qquad
\mathrm{IoU}\!\left(\mathcal{B}_c(\mathbf{p}_c), \textstyle\bigcup_{j \in \mathcal{P}_p} \mathcal{B}_j\right)=0,
\end{equation}
where $\mathcal{H}_p$ is the parent's support region, $\mathcal{B}_c$ is the projected child footprint, and $\mathcal{P}_p$ contains siblings already placed on $p$.
The support predicate prevents unstable placement, while the IoU term avoids inter-object collisions.
Manipulated objects must additionally lie within the robot's reachable, forward-facing interaction region; when no feasible pose is found, the system resamples the candidate or invokes a relation-specific fallback.
After placement, we settle movable objects under gravity in SAPIEN~\cite{Xiang_2020_SAPIEN} to resolve residual penetrations and floating artifacts.
We then export the stabilized 6-DoF poses to a standardized layout configuration that can be loaded across simulators for visualization and batched policy training.
RoboVerse~\cite{geng2025roboverse} provides an online example of directly loading this standardized layout description across different simulators (Fig.~\ref{fig:cross_sim}).\footnote{\url{https://roboverse.wiki/metasim/get_started/quick_start/16_embodiedgen_layout}}

\subsection{Large-Scale Scenes Generation}
\label{sec:large_scale_scene_generation}

\begin{figure}[t]
  \centering
  \includegraphics[width=1.0\linewidth]{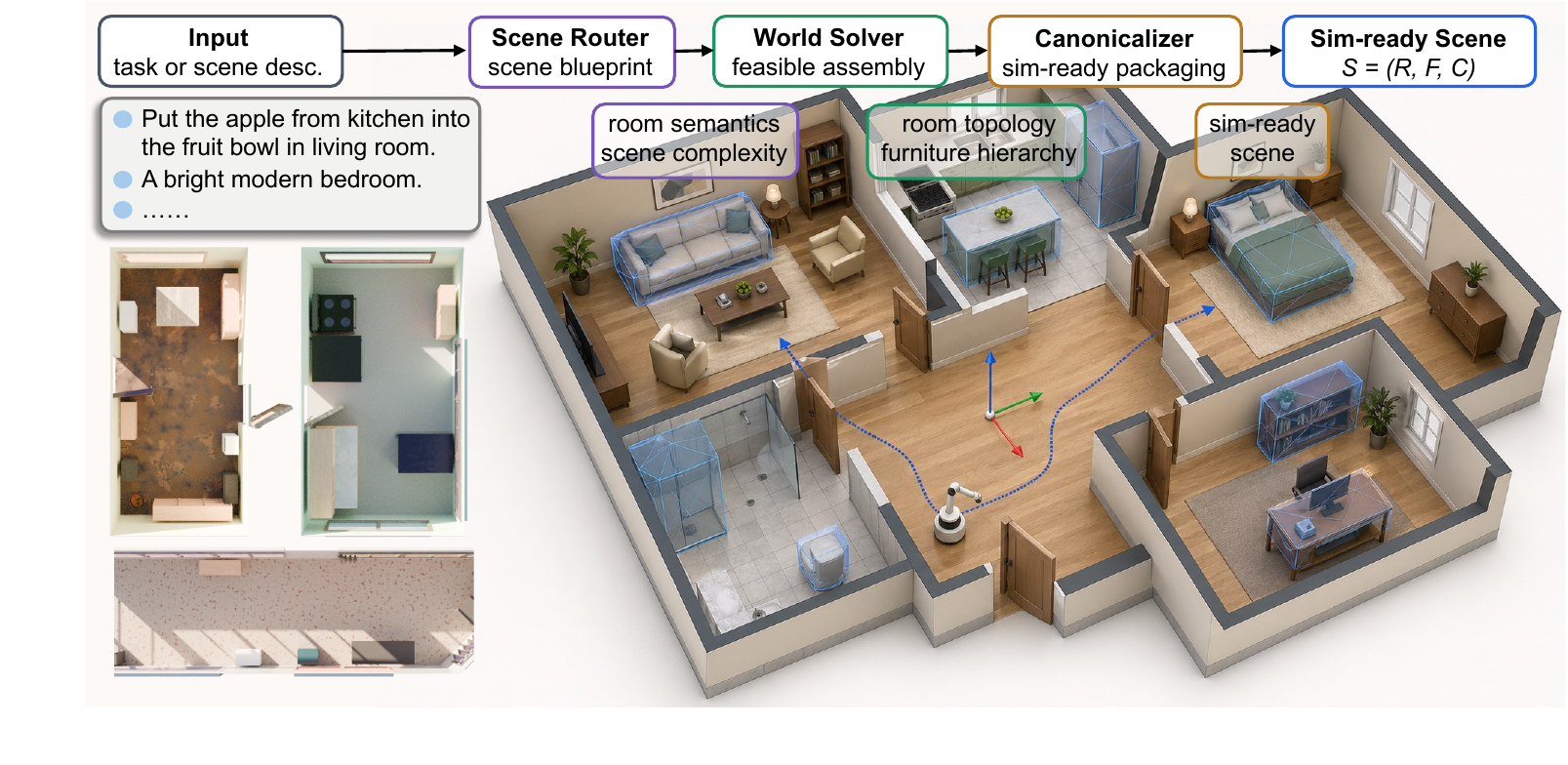}
  \caption{
    Large-scale scenes generation. A task description is first distilled into a
    scene blueprint, then assembled under spatial constraints as a feasible
    multi-room assembly, and finally canonicalized into a sim-ready background
    that can serve as the \textsc{Background} node for task-driven foreground
    layouts.
  }
  \label{fig:large_scale_scene_gen}
\end{figure}

\textbf{Problem definition.}
The task-driven interactive worlds generation module in Sec.~\ref{sec:layout_generation} explicitly factors out the \emph{task-relevant interactive foreground assets} from the world, leaving an abstract interface occupied by a placeholder background node.
This section answers the other half of that interface: how to automatically synthesize, behind that placeholder, a \emph{large-scale, multi-room, navigable} simulation-ready indoor scene, so that embodied agents are no longer confined to tabletop-scale geometry but can perform long-horizon navigation and mobile manipulation within a unified physical world.
Formally, given a task description $\mathcal{T}$, this module outputs a triple $\mathcal{S} = (\mathcal{R}, \mathcal{F}, \mathcal{C})$, where $\mathcal{R}$ is a room topology graph annotated with door and window connections, $\mathcal{F}$ is a per-room, individually addressable set of furniture instances (each carrying a visual mesh, a collision proxy, and physical parameters), and $\mathcal{C}$ is a globally consistent house-level coordinate frame.
Compared to the panorama-back-projected single-mesh background of EmbodiedGen V1~\cite{wang2025embodiedgengenerative3dworld}, $\mathcal{S}$ simultaneously exposes a real room topology, traversable openings, and independently editable furniture entities---the conditions such tasks require to be solvable in simulation.

\textbf{Pipeline overview.}
We propose a three-stage pipeline that explicitly decouples natural-language-driven semantic reasoning from deterministic geometric constraint solving:
\textbf{(i)~Task-conditioned routing} encodes $\mathcal{T}$ into a small set of discrete control signals that specify which room semantics to instantiate and at what target complexity;
\textbf{(ii)~Hierarchical scene solving} reshapes a procedural indoor generation framework~\cite{infinigen2024indoors} from a render-oriented synthesizer into a simulation-oriented solver, producing room- or house-level geometry populated with a furniture skeleton;
\textbf{(iii)~Simulator-agnostic canonicalization} performs per-instance decomposition, collision-proxy generation, and coordinate normalization, then delivers the scene through URDF as a standardized intermediate representation, reusing the unified format converter of Sec.~\ref{sec:sim_ready_asset} to reach all major downstream simulators.
This shifts the burden of spatial consistency onto a constrained solver, complementing rather than replacing LLM-driven layout work~\cite{yang2024holodeck,feng2023layoutgpt}: the language model makes only discrete semantic decisions, while the solver guarantees geometric and topological feasibility.

\textbf{Task-conditioned routing.}
A vision-language model maps the task $\mathcal{T}$ to two discrete controls: room scope and scene complexity.
Local tasks select a plausible room category, whereas cross-room or long-horizon tasks trigger a whole-house joint solve.
The complexity level $\ell \in \{\textsc{Minimalist}, \textsc{Simple}, \textsc{Medium}, \textsc{Detail}\}$ controls furniture and clutter density, providing an interpretable interface between task requirements and solver cost.

\textbf{Hierarchical scene solving.}
The solver places three semantic scales of furniture in a coarse-to-fine order under floor-plan and cross-room traversability constraints: it first solves the placement and orientation of skeleton-level furniture (large items such as beds, sofas, and cabinets that define a room's function), then mid-scale objects on top of their supporting surfaces, and finally tabletop-scale clutter.
The complexity tier $\ell$ controls how many of these levels the solver activates, exposing scene density as a single tunable axis so that generation cost scales with task difficulty, from near-empty rooms to fully decorated interiors.
To make the output truly usable for physical simulation rather than offline rendering, we reshape the procedural generator from \emph{render-oriented} to \emph{simulation-oriented}: we suppress geometry that is unparseable to physics back-ends or merely decorative, and reallocate the solver's budget toward maintaining feasibility on large-scale multi-room scenes.
The solver resamples infeasible samples under room-connectivity and navigable-opening constraints.

\textbf{Simulator-agnostic canonicalization.}
The geometry produced by the solver is still in a render-oriented representation and must undergo a sim-ready packaging step before it is consumable by physics engines.
We extend the scene-level export with three components:
\textbf{(i)~Per-instance decomposition} splits the house-level geometry along furniture and architectural units into individually loadable, individually replaceable instance entities, so that the background generated here can be seamlessly mounted as a \textsc{Background} node of the Scene Graph in Sec.~\ref{sec:layout_generation}, with its furniture instances replaceable or extensible by foreground objects on demand;
\textbf{(ii)~Convex collision proxy} batch-applies the same CoACD~\cite{wei2022coacd} convex decomposition used in Sec.~\ref{sec:sim_ready_asset} across all furniture instances, replacing visual meshes with compact convex hulls as collision proxies and avoiding the contact instability and performance degradation typical of non-convex meshes at scene scale;
\textbf{(iii)~Scene-level canonicalization} aligns the centroid of the house-level geometry to the world origin, eliminating the global pose drift that otherwise appears across random seeds and ensuring comparability for downstream policy training under bulk data generation.
The packaging stage natively produces both URDF and USD and reuses the unified format converter of Sec.~\ref{sec:sim_ready_asset}, so house-level backgrounds and object-level assets reach all major simulators through the same delivery path, with consistent physical semantics and contact behavior across back-ends.
This section therefore delivers not a non-editable visual shell but a \emph{routable, addressable, replaceable, and engine-portable structured sim-ready background}.
Together with the object-level assets of Sec.~\ref{sec:sim_ready_asset} and the task-level layouts of Sec.~\ref{sec:layout_generation}, it forms the unified world representation of \modelname: usable on its own for navigation and multi-room exploration, or composed as the physical background of foreground layouts to constitute end-to-end embodied simulation worlds.

\subsection{Vibe Coding for Stateful Sim-Ready 3D World Editing}
\label{sec:vibe_coding_engine}

\textbf{Motivation.}
We use the term \emph{Vibe Coding} for iteratively generating and editing simulation-ready 3D worlds through natural-language dialogue: the user expresses intent conversationally while deterministic, physics-aware skill backends enforce feasibility and sim-ready output contracts---much as a developer \emph{vibes} with an AI coding assistant while the compiler enforces type correctness.
Such authoring is intrinsically iterative---refining instances, adjusting spatial relations, restyling assets, and validating physics---yet neither conventional 3D pipelines (modeling, physics annotation, format-specific export) nor prompt-to-scene generators, which regenerate the whole scene on each prompt, support state-preserving local edits.
The generators of Sec.~\ref{sec:sim_ready_asset}--\ref{sec:large_scale_scene_generation} are likewise single-shot, and modern LLM agents, though able to invoke domain skills through typed tool calls, lack a self-describing skill suite for sim-ready 3D worlds.
We therefore organize these modules under an \emph{agent--skill--harness} abstraction: the solvers become callable sim-ready skills, and a shared harness maintains a world state that evolves across dialogue turns, converting single-shot generation into an online editing kernel that reuses the backbone of Eq.~\eqref{eq:placement}, CoACD collision proxies, and physics settling.

\textbf{Architecture and world state.}
The editing engine contains three components.
(i)~The \emph{agent} is an LLM-based coordinator responsible for dialogue understanding, intent parsing, skill selection, argument completion, and feedback explanation.
(ii)~The \emph{skills} are self-contained capability units; each skill exposes a natural-language description of its usage, inputs, outputs, and failure modes, and is backed by deterministic generators, solvers, or exporters from Sec.~\ref{sec:sim_ready_asset}--\ref{sec:large_scale_scene_generation}.
(iii)~The \emph{harness} is the runtime layer that bridges the agent and skills, maintaining the skill registry, dispatch logic, shared world state, failure loop, and edit log.
Given a continuing dialogue stream $\{u_1, u_2, \dots, u_t\}$, the harness maintains an evolving simulation-ready world state
\begin{equation}
\label{eq:vibe_state}
\mathcal{S}_t = \big(\mathcal{G}_t,\, \mathcal{A}_t,\, \mathcal{P}_t,\, \mathcal{H}_t\big),
\end{equation}
where $\mathcal{G}_t$ is the typed Scene Graph, $\mathcal{A}_t$ the sim-ready assets, $\mathcal{P}_t$ their 6-DoF poses, and $\mathcal{H}_t$ the dialogue and skill-invocation history.
For each instruction, the agent selects a skill and the harness validates its arguments, executes the deterministic backend, and commits a bounded state update.
Every update preserves geometric and physical feasibility, records an auditable edit in $\mathcal{H}_t$, and keeps assets and poses consistent across simulator backends.

\begin{figure}[H]
  \centering
  \includegraphics[width=0.98\linewidth]{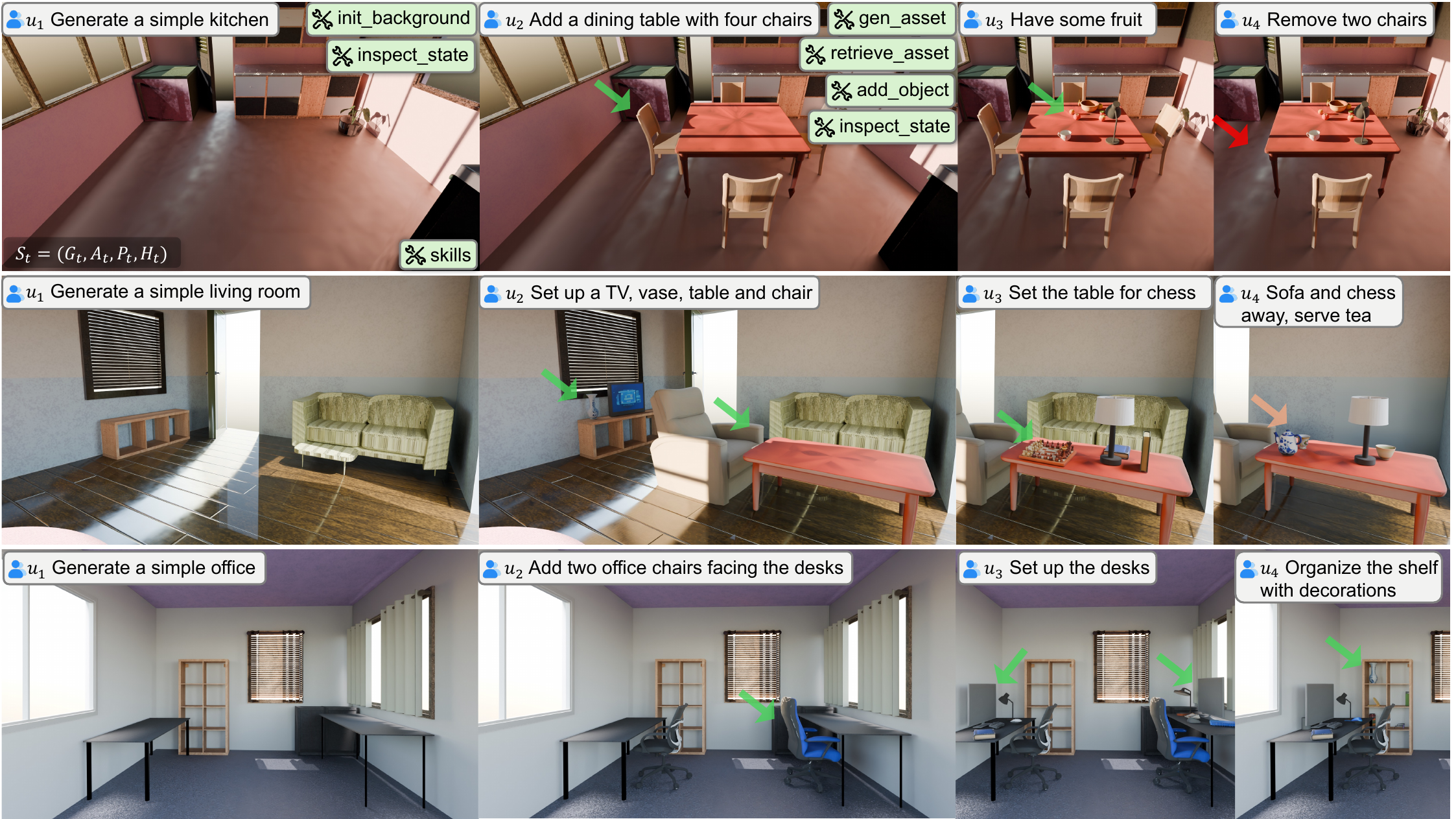}
  \caption{
    Three Vibe Coding 3D editing sessions (kitchen, top; living room, middle;
    office, bottom). From a bare background $\mathcal{S}_0$, each session
    proceeds through three natural-language turns in which typed skills commit
    bounded deltas $\Delta\mathcal{S}$ to
    $(\mathcal{G},\mathcal{A},\mathcal{P},\mathcal{H})$. Green and red arrows
    indicate added and removed instances; all sessions end as sim-ready,
    exportable scenes.
  }
  \label{fig:vibe_coding_engine}
\end{figure}

\textbf{Skill suite and runtime contract.}
We organize the skill suite around four abstractions required by stateful sim-ready world generation: asset grounding, world composition, stateful editing, and execution validation (Table~\ref{tab:capability_primitives}).
Each skill declares its trigger, typed arguments, outputs, and failure behavior.
After \textsc{Parse} and \textsc{Ground}, the harness validates the arguments, invokes the deterministic backend, and commits the resulting bounded $\Delta\mathcal{S}$ in the shared asset-and-layout representation.
Failed calls return structured diagnostics for retry, disambiguation, or fallback without modifying the world state.
The harness is not tied to a particular agent implementation: our reference adapter integrates with OpenAI Codex~\cite{openai2025codex} and Gemini CLI~\cite{google2025geminicli} through a shared plugin layer, and any agent framework supporting typed tool calls or slash-command plugins can connect to the same skill set.

\begin{table}[ht]
\centering
\footnotesize
\caption{Core skill abstractions exposed by the Vibe Coding agent--skill harness.}
\label{tab:capability_primitives}
\setlength{\tabcolsep}{4pt}
\renewcommand{\arraystretch}{1.12}
\begin{tabularx}{\linewidth}{@{}>{\raggedright\arraybackslash}p{2.55cm}
                              >{\raggedright\arraybackslash}p{3.35cm}
                              X@{}}
\toprule
\textbf{Abstraction} & \textbf{Skill} & \textbf{Role in the world-state transition} \\
\midrule
\multirow{4}{=}{Asset grounding}
 & \texttt{asset-creator}    & Materializes open-vocabulary object intent into sim-ready asset candidates. \\
 & \texttt{asset-retrieval}  & Grounds object references to reusable assets when generation is unnecessary. \\
 & \texttt{asset-process}    & Preserves metric and geometric consistency under asset-level transformations. \\
 & \texttt{asset-converter}  & Projects canonical asset representations into simulator-specific formats. \\
\midrule
\multirow{3}{=}{World composition}
 & \texttt{background-creator} & Synthesizes task-compatible background context for interactive layouts. \\
 & \texttt{room-creator}     & Produces structured room- and house-level worlds with canonical scene entities. \\
 & \texttt{layout-creator}   & Instantiates task semantics as physically grounded foreground--background layouts. \\
\midrule
\shortstack[l]{Stateful\\editing}
 & \texttt{spatial-computing} & Commits bounded scene edits by grounding language to addressable instances and collision-aware spatial constraints. \\
\midrule
\shortstack[l]{Execution\\validation}
 & \texttt{sim-runner}       & Closes the loop by executing the current world state in simulation and returning visual or policy-relevant feedback. \\
\bottomrule
\end{tabularx}
\end{table}

\textbf{Agent--skill editing loop.}
Algorithm~\ref{alg:vibe_loop} summarizes the \textsc{Parse}--\textsc{Ground}--\textsc{Invoke}--\textsc{Commit} loop.
Each instruction produces a bounded state delta under the current feasibility constraints; successful calls commit and render the new state, while failed calls return diagnostics without changing it.

\begin{algorithm}[H]
\caption{Agent--Skill Interactive Editing Loop.}
\label{alg:vibe_loop}
\begin{algorithmic}[1]
\Require dialogue stream $\{u_t\}$, initial world state $\mathcal{S}_0$
\For{each instruction $u_t$}
    \State $(\omega,\, \alpha_{\mathrm{NL}}) \gets \textsc{Parse}(u_t,\, \mathcal{S}_t)$ \Comment{select skill and NL arguments}
    \State $\alpha \gets \textsc{Ground}(\alpha_{\mathrm{NL}},\, \mathcal{S}_t)$ \Comment{resolve typed world references}
    \State $\Delta\mathcal{S} \gets \textsc{Invoke}\big(\omega,\, \alpha,\, \mathcal{C}(\mathcal{S}_t)\big)$ \Comment{execute under constraints}
    \If{$\Delta\mathcal{S} = \bot$}
        \State $\textsc{Diagnose}(\omega,\, \alpha,\, \mathcal{S}_t)$ \Comment{return diagnostics; no state mutation}
        \State \textbf{continue}
    \EndIf
    \State $\mathcal{S}_{t+1} \gets \textsc{Commit}(\mathcal{S}_t,\, \Delta\mathcal{S})$ \Comment{atomic state update}
    \State $\textsc{Render}(\mathcal{S}_{t+1})$ \Comment{refresh simulation preview}
\EndFor
\end{algorithmic}
\end{algorithm}

\begin{figure}[t]
  \centering
  \includegraphics[width=1.0\linewidth]{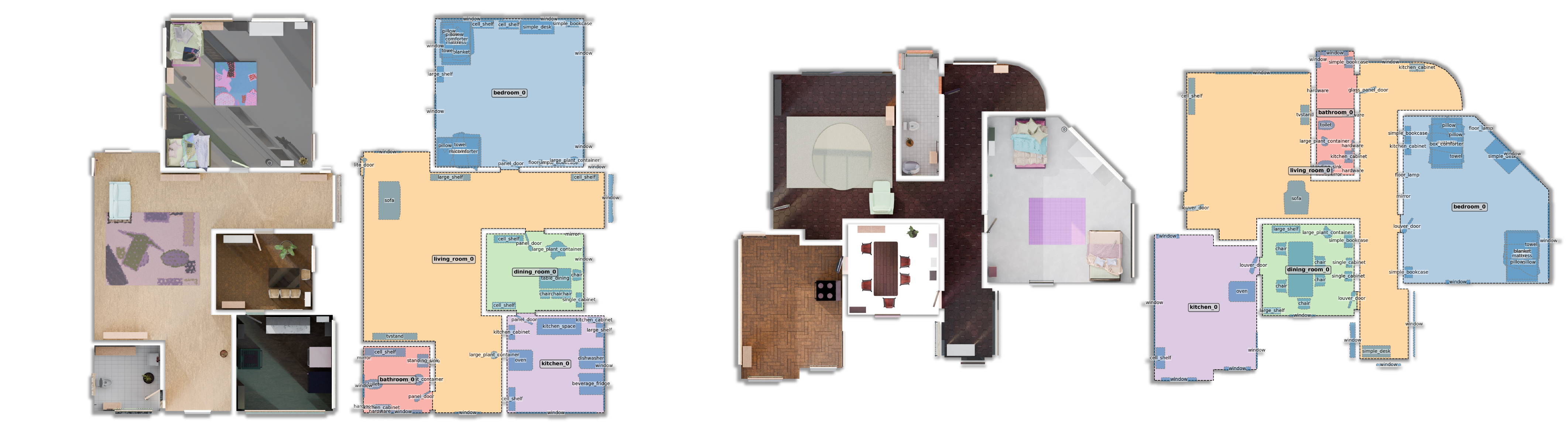}
  \caption{
    Floorplan canvas of the \texttt{spatial-computing} skill. Each pair shows a
    top-down rendering (left) and the corresponding symbolic floorplan with
    room and instance labels (right). Open-vocabulary references are grounded
    against this canvas, and the skill evaluates Eq.~\eqref{eq:placement} on
    its room polygons and instance bounding boxes.
  }
  \label{fig:floorplan_canvas}
\end{figure}

\textbf{Instance grounding and spatial editing.}
\textsc{Ground}, handled by the agent, is the interface between open-ended language and the symbolic world state.
It resolves category references (``the chair''), attribute references (``the largest piece of furniture''), and historical anaphora (``the apple I just placed'') against the current scene graph, instance appearance, spatial coordinates, and recent edits in $\mathcal{H}_t$, producing typed arguments $\alpha$ for the selected skill $\omega$; low-confidence cases return top-$k$ candidates for user disambiguation.
For spatial edits, \textsc{Ground} maps references to an \texttt{instance\_key} and \texttt{room\_id} and dispatches them to the \texttt{spatial-computing} skill, which exposes the scene as a room-partitioned 2D floorplan of addressable instances (Fig.~\ref{fig:floorplan_canvas}) and resolves the \texttt{ON}/\texttt{BESIDE}/\texttt{IN} relations by reusing the collision-IoU term of Eq.~\eqref{eq:placement}, with its support test generalized from object top-surfaces to room free-floor polygons.
The offline placement solver thus becomes the online core of this skill, and every committed edit keeps $\mathcal{G}_t$ and $\mathcal{P}_t$ consistent across downstream simulators.

\FloatBarrier
\section{Experiments}
\label{sec:experiments}

\subsection{Sim-Ready Pipeline Quality Evaluation}
\label{sec:sim_ready_evaluation}

We ablate each stage of the sim-ready pipeline on 200 held-out assets spanning diverse object categories.
The default \emph{Full pipeline} includes mesh fixing, convex decomposition, and the hierarchical quality checker; each variant removes one component while keeping the rest unchanged.
For this evaluation, we use SAM3D~\cite{sam3dteam2025sam3d3dfyimages} as the image-to-3D model, and run all experiments on a single NVIDIA RTX 4090 GPU.

\begin{table}[htbp]
\centering
\caption{Ablation study on 200 held-out assets. Each row removes one component from the full pipeline. Time and mesh sizes are reported as mean $\pm$ std. Best results are in \textbf{bold}.}
\label{tab:sim_ready_ablation}
\small
\setlength{\tabcolsep}{3pt}
\begin{tabularx}{\linewidth}{l *{5}{>{\centering\arraybackslash}X}}
\toprule
Setting & \shortstack{Human\\Accept. $\uparrow$} & \shortstack{Collision\\Success $\uparrow$} & \shortstack{Time\\(min) $\downarrow$} & \shortstack{Visual\\Mesh (MB) $\downarrow$} & \shortstack{Collision\\Mesh (MB) $\downarrow$} \\
\midrule
Full pipeline          & \textbf{96.5\%} & \textbf{98.6\%} & 2.6 $\pm$ 0.4 & \textbf{1.43 $\pm$ 0.63} & \textbf{0.29 $\pm$ 0.21} \\
w/o Quality checker    & 91.0\% & 98.1\% & \textbf{2.2 $\pm$ 0.4} & 1.44 $\pm$ 0.63 & 0.30 $\pm$ 0.22 \\
w/o Mesh fixing        & 95.5\% & 98.3\% & 21.3 $\pm$ 22.8 & 51.63 $\pm$ 25.87 & 0.31 $\pm$ 0.26 \\
w/o Convex decomp.     & 94.5\% & 96.5\% & 2.3 $\pm$ 0.3 & 1.45 $\pm$ 0.64 & 1.45 $\pm$ 0.64 \\
\bottomrule
\end{tabularx}
\end{table}

Table~\ref{tab:sim_ready_ablation} reports five metrics.
\emph{Human Acceptance} is the fraction of assets judged sim-ready by annotators, considering consistency with the input condition, geometric plausibility, coherence of invisible surfaces, and overall usability as a simulation asset.
\emph{Collision Success} is the average success rate of scripted Franka Panda top-down grasp-and-lift trials in SAPIEN~\cite{Xiang_2020_SAPIEN}: we run 4 trials per asset at evenly spaced yaw angles and count a trial as successful if the object is lifted above an adaptive height threshold proportional to its bounding-box height.
The remaining columns report per-asset processing time and exported mesh sizes.

\textbf{Quality checker.}
Removing the quality checker lowers Human Acceptance by 5.5 points (96.5\%\,$\to$\,91.0\%) while saving only 0.4\,min per asset.
Without the generate--verify--retry loop, defective samples pass through uncorrected, degrading visual and semantic completeness.
Mesh sizes and Collision Success remain largely unchanged, confirming that the checker targets perceptual defects rather than geometric properties.

\textbf{Mesh fixing.}
Without mesh fixing, downstream processing becomes substantially less efficient: per-asset runtime increases by approximately $8\times$, from $2.6 \pm 0.4$ to $21.3 \pm 22.8$\,min, and the visual mesh size grows from 1.43\,MB to 51.63\,MB.
This degradation occurs because raw generative outputs contain redundant faces and topological defects that severely slow UV unwrapping, texture baking, and convex decomposition~\cite{wei2022coacd}; meshes exceeding 50\,MB per object are also impractical for simulators that batch-load hundreds of assets.

\textbf{Convex decomposition.}
Without convex decomposition the collision mesh reverts to the full visual mesh as the collision proxy (0.29\,$\to$\,1.45\,MB), forcing the simulator to operate on non-convex surfaces.
This reduces Collision Success from 98.6\% to 96.5\% due to unstable contact and grasp failure, while adding negligible runtime overhead (2.6 vs.\ 2.3\,min).
Although the absolute drop is modest, such contact errors can accumulate in long-horizon embodied manipulation pipelines.

Overall, the three components address complementary failure modes---perceptual acceptance, deployment efficiency, and contact reliability---and together yield 96.5\% Human Acceptance and 98.6\% Collision Success at 2.6\,min per asset.

\subsection{Affordance Pipeline Quality Evaluation}
\label{sec:affordance_evaluation}

We evaluate the affordance autolabeling pipeline on 200 sim-ready assets spanning diverse object categories under three matched settings.
The \emph{Baseline} uses raw P3-SAM part segmentation~\cite{ma2025p3sam}, part-wise semantic annotation, and grasp generation; the two variants progressively add geometry-consistent segmentation post-processing and VLM-guided part merging.
All variants use the same asset set and a cascaded protocol, in which each stage is measured only on assets that pass the preceding stage.
We run the evaluation on a single NVIDIA RTX 4090 GPU and instantiate the VLM-assisted verification and merging module with GPT-5.4~\cite{openai2026gpt5dot4}.
For grasp validation, each retained grasp is executed in SAPIEN, and we discard it if the object-to-gripper slip exceeds $5$\,cm or $30^\circ$ during the validation sequence.

\begin{table}[htbp]
\centering
\small
\caption{\textbf{Ablation study of the affordance autolabeling pipeline on 200 assets.}
Stage pass rates are conditional on the previous successful stage, while the end-to-end pass rate is the product of part segmentation, part semantic annotation, and object-level grasp coverage. 
Runtime is reported per asset as mean $\pm$ std over both successful and failed assets.}
\label{tab:affordance_ablation}
\setlength{\tabcolsep}{1pt}
\begin{tabularx}{\linewidth}{@{}l *{5}{>{\centering\arraybackslash}X}@{}}
\toprule
Setting & \shortstack{Segmentation\\Pass Rate $\uparrow$} & \shortstack{Semantic\\Validity Rate $\uparrow$} & \shortstack{Grasp\\Coverage Rate $\uparrow$} & \shortstack{Affordance\\Pass Rate $\uparrow$} & \shortstack{Runtime\\(s) $\downarrow$} \\
\midrule
Baseline                  & 47.0\% & 98.9\% & 66.7\% & 31.0\% & 109 $\pm$ 45 \\
+ Post-process            & 56.5\% & 97.3\% & \textbf{74.6\%} & 41.0\% & 105 $\pm$ 41 \\
+ Post-process + VLM merging      & \textbf{69.5\%} & \textbf{99.3\%} & 72.5\% & \textbf{50.0\%} & \textbf{94 $\pm$ 30} \\
\bottomrule
\end{tabularx}
\end{table}

Table~\ref{tab:affordance_ablation} reports stage-wise and end-to-end metrics.
\emph{Segmentation Pass Rate} measures whether the VLM checker accepts the decomposition as physically meaningful functional parts after inspecting the aligned RGB and mask grids, and whether the face-level masks are accurate enough for downstream labeling.
\emph{Semantic Validity Rate} measures whether each accepted part receives a valid structured annotation with semantic name, graspability flag, grasp scenarios, functional labels, and appearance description after the checker-repair loop.
\emph{Grasp Coverage Rate} is computed on the semantic-passed assets and requires the whole object to have at least one simulation-validated candidate grasp pose in the final affordance annotation, while assets that are intrinsically unsuitable for mechanical parallel-jaw grasping and lifting, such as large household appliances, are treated as satisfying this criterion.
\emph{Segmentation Pass Rate} and \emph{Semantic Validity Rate} are first screened by the VLM checker and then manually cross-validated, whereas \emph{Grasp Coverage Rate} is computed automatically from the final affordance annotations.
Under this cascaded evaluation, \emph{Affordance Pass Rate} is computed as the product of the three stage-wise rates and represents the end-to-end yield of assets with valid part-level semantics and either at least one physically executable grasp or an explicit graspability exemption.

\textbf{Post-processing.}
Adding geometry-consistent post-processing improves \emph{Segmentation Pass Rate} from 47.0\% to 56.5\%, increasing the fraction of assets with usable face-level part masks.
This improvement reflects the role of smooth-component merging and surrounded-fragment relabeling, which reduce projection-induced boundary noise and isolated part fragments without collapsing genuine geometric discontinuities.
The \emph{Semantic Validity Rate} remains high on the segmentation-passed assets (98.9\%\,$\to$\,97.3\%), while the conditional \emph{Grasp Coverage Rate} increases from 66.7\% to 74.6\%.
These improvements in segmentation quality and grasp coverage propagate through the cascade, raising \emph{Affordance Pass Rate} from 31.0\% to 41.0\%.

\textbf{VLM verification.}
Adding VLM-based verification further raises \emph{Segmentation Pass Rate} to 69.5\%, substantially increasing the pool of assets that proceed to grasp generation.
By merging regions that represent the same functional part but are separated in the raw segmentation, the VLM checker produces part decompositions that better align with actionable object structure.
The \emph{Semantic Validity Rate} increases to 99.3\%, while the conditional \emph{Grasp Coverage Rate} remains competitive at 72.5\%, suggesting that the VLM module primarily increases the number of reliable part-level carriers while preserving strong grasp coverage.
At the end-to-end level, the full configuration achieves a 50.0\% \emph{Affordance Pass Rate}, improving over the baseline by 19.0 percentage points.

\textbf{Efficiency.}
All variants run in roughly 1.6--1.8 minutes per asset.
The full variant is the fastest in this measurement ($94.0 \pm 30.2$\,s), compared with $108.5 \pm 45.1$\,s for the baseline.
Post-processing and VLM verification add cost to part segmentation because they require face-label refinement, checker rendering, and up to three merge-and-recheck attempts.
However, by suppressing redundant or over-fragmented parts, these components reduce the downstream workload for part-wise semantic annotation, grasp generation, and simulation-based grasp validation; for example, VLM verification reduces the average number of parts from 5.3 to 3.6.
This reduction in downstream computation outweighs the added segmentation overhead, leading to a lower overall per-asset runtime.

\FloatBarrier
\subsection{Task-Driven Interactive Worlds Generation Evaluation}
\label{sec:layout_evaluation}

\textbf{Experimental setup.}
We generate 150 diverse natural-language manipulation tasks covering different indoor backgrounds, contexts, manipulated objects, and distractors.
Each task is generated end-to-end following Sec.~\ref{sec:layout_generation}: an LLM first parses the task into a Scene Graph, the system then generates the background and required sim-ready object assets online, and BFS-based spatial placement with SAPIEN physics settling produces a directly loadable interactive world layout.
Unless otherwise stated, all efficiency numbers are measured under fully online sequential generation on a single NVIDIA RTX 4090 GPU.
We manually inspect the final interactive worlds for direct usability in downstream simulation, using criteria of task-semantic consistency, correct spatial relations, reasonable object scale, physical stability, and robot executability.

Table~\ref{tab:layout_pipe} reports evaluation results from four perspectives: task-to-graph generation, asset instantiation, automated quality checking, and final world-level acceptance.
The 150 generated interactive worlds contain 778 sim-ready object asset instances across 128 object categories, with an average of 5.19 interactive assets per world.
These results indicate that the Scene Graph representation stably decomposes open-ended tasks while maintaining broad world-composition diversity.

\begin{table}[htbp]
\centering
\small
\caption{\textbf{Stage-wise profiling of the task-driven interactive worlds generation pipeline (Sec.~\ref{sec:layout_generation}).}
Results are aggregated over 150 generated interactive worlds under sequential execution on a single NVIDIA RTX 4090 GPU.
QA pass rates denote single-attempt pass rates across asset generation attempts.}
\label{tab:layout_pipe}
\begin{tabular}{@{}llc@{}}
\toprule
\textbf{Category} & \textbf{Metric} & \textbf{Value} \\
\midrule
\textbf{Task-to-Graph}
  & Generated task-conditioned worlds         & 150 worlds \\
  & Avg.\ interactive asset instances per world & 5.19 instances \\
  & Distinct object categories covered        & 128 categories \\
\midrule
\textbf{Asset Instantiation}
  & Background asset instances generated     & 150 instances \\
  & Object asset instances generated         & 778 instances \\
  & Time per background asset                & $25.5 \pm 3.5$ min \\
  & Time per object asset                    & $3.6 \pm 1.1$ min \\
\midrule
\textbf{Asset QA}
  & Semantic Appearance                       & 76.2\% \\
  & Mesh Geometry                             & 75.9\% \\
  & Cross-modal Text-to-3D Alignment          & 91.0\% \\
  & Avg.\ attempts per valid asset            & 1.35 \\
\midrule
\textbf{World-Level Outcome}
  & Total time per world                      & $47.7 \pm 5.4$ min \\
  & Final environment acceptance rate         & 83.3\% \\
\bottomrule
\end{tabular}
\end{table}

\FloatBarrier

\begin{figure}[!htbp]
  \centering
  \includegraphics[width=0.99\linewidth]{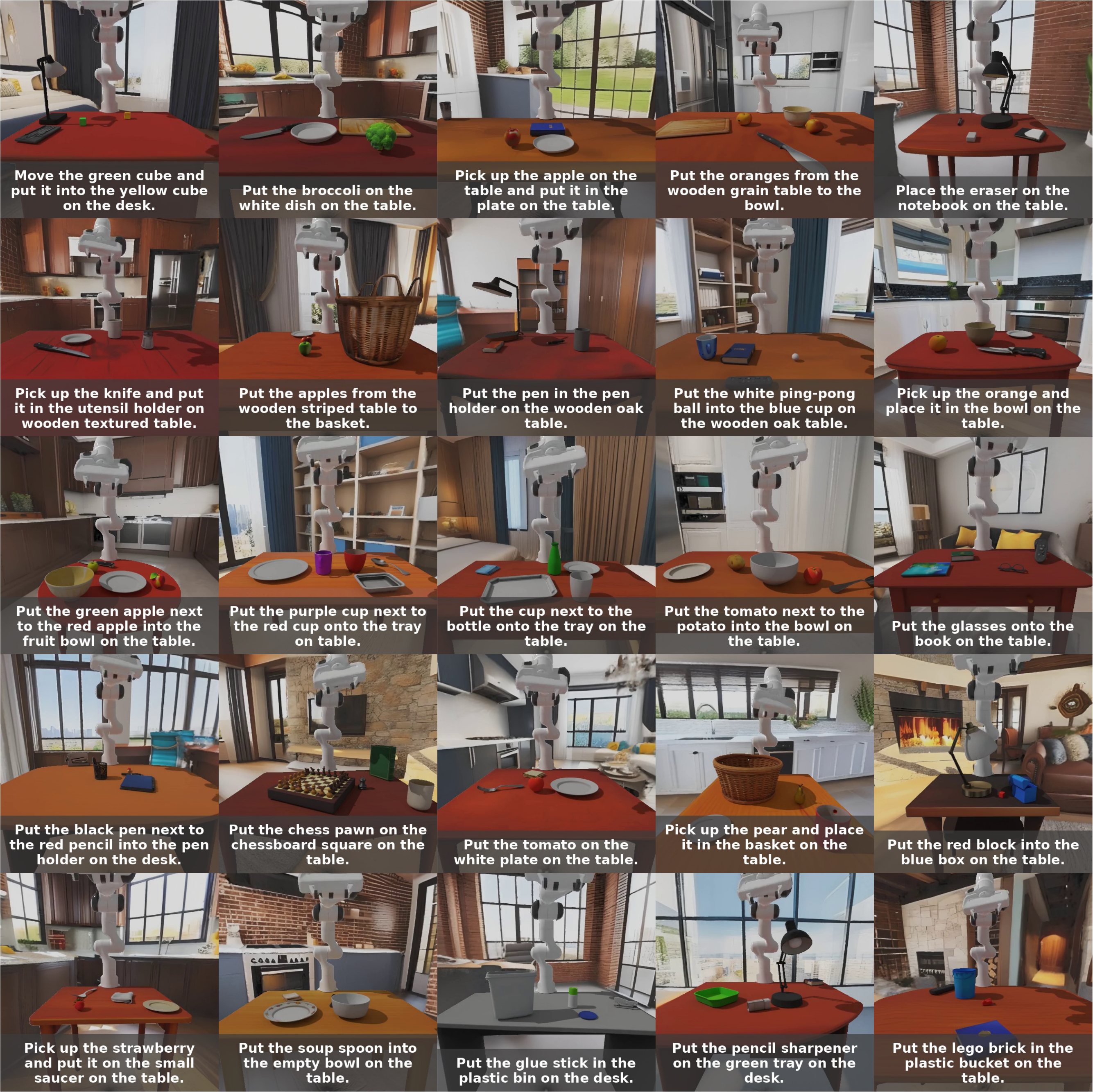}
  \caption{
    Qualitative examples of task-driven interactive worlds generation. Each world is
    generated from a task description via a task-conditioned Scene Graph, and
    contains a background, contexts, manipulated objects, and distractors
    arranged into a simulation-ready layout.
  }
  \label{fig:layout_demo}
\end{figure}

Fig.~\ref{fig:layout_demo} shows representative layouts generated by the task-driven interactive worlds generation pipeline.

Beyond these qualitative examples, the stage-wise profiling in Table~\ref{tab:layout_pipe} shows that fully online generation takes $47.7 \pm 5.4$ minutes per world, with background synthesis being the dominant cost ($25.5 \pm 3.5$\,min).
Each object asset takes $3.6 \pm 1.1$\,min to generate.
Since the Scene Graph explicitly separates backgrounds from interactive assets, the system can reuse existing background or object instances from an offline asset library in practical use, avoiding repeated execution of the most expensive generation steps and reducing world generation to the order of minutes.

\begin{figure}[t]
  \centering
  \includegraphics[width=0.78\linewidth]{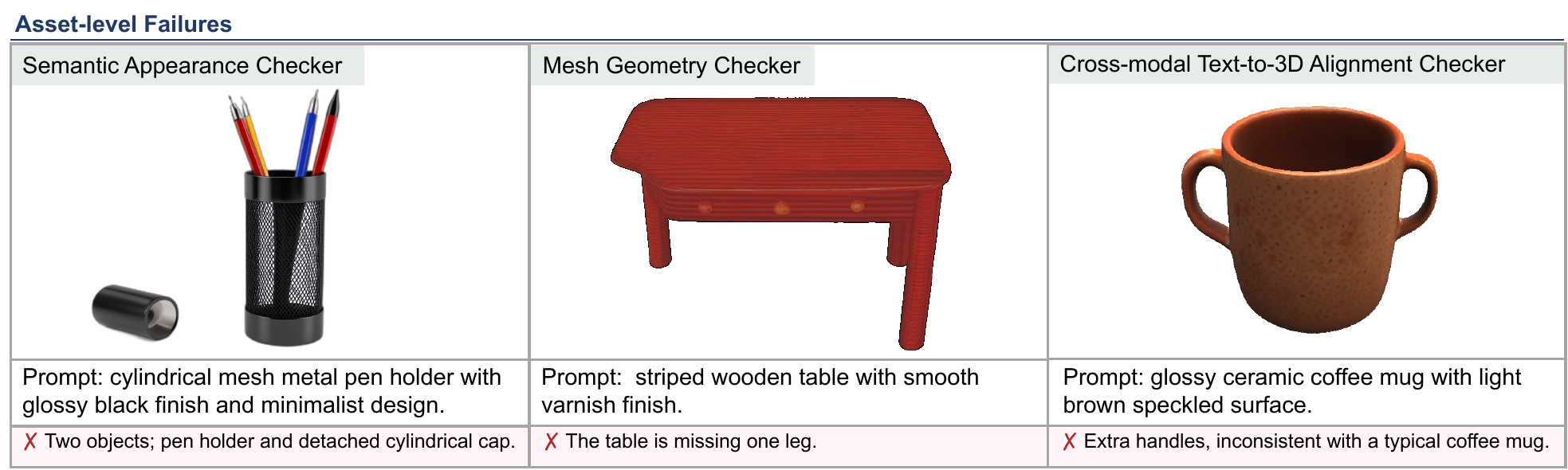}
  \vspace{0.3em}
  \includegraphics[width=0.78\linewidth]{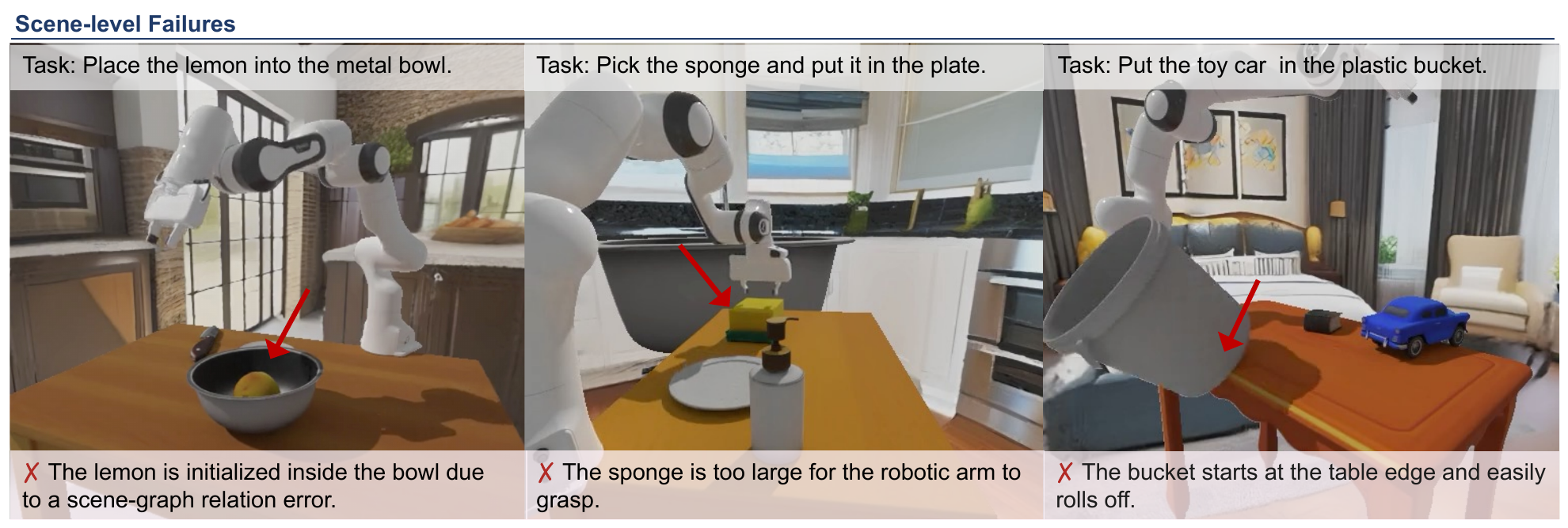}
  \caption{
    Representative failure cases in task-driven interactive worlds generation. Top:
    asset-level failures filtered by the automated QA modules, including
    semantic appearance mismatch, mesh defects, and text-to-3D drift. Bottom:
    major world-level failure cases corresponding to the non-accepted portion
    of the final environment inspection, including task-constraint violation,
    object-scale mismatch and unstable placement.
  }
  \label{fig:layout_failure_cases}
\end{figure}

Automated quality checks mainly operate during asset instantiation and prevent errors from propagating across intermediate representations.
The Semantic Appearance checker verifies whether the foreground image matches the target category and key visual attributes; the Mesh Geometry checker checks whether the generated mesh is complete and free of major geometric defects; and the Cross-modal Text-to-3D Alignment checker verifies whether the final 3D asset remains semantically consistent with the original text description, capturing semantic drift introduced during 3D generation.
The three checkers achieve single-attempt pass rates of 76.2\%, 75.9\%, and 91.0\%, respectively.
With the generate--verify--retry mechanism, each valid asset requires only 1.35 generation attempts on average.
Manual inspection shows that 83.3\% of final interactive worlds can be used for downstream simulation without manual modification.
As shown in Fig.~\ref{fig:layout_failure_cases}, the remaining failures mainly arise from object-scale mismatch, local geometry defects, or imperfect initial spatial placement, and can typically be corrected by resampling or minor manual adjustment.

\FloatBarrier

\subsection{Downstream Closed-Loop Validation}
\label{sec:closed_loop_validation}

\begin{figure}[!htbp]
  \centering
  \includegraphics[width=\linewidth]{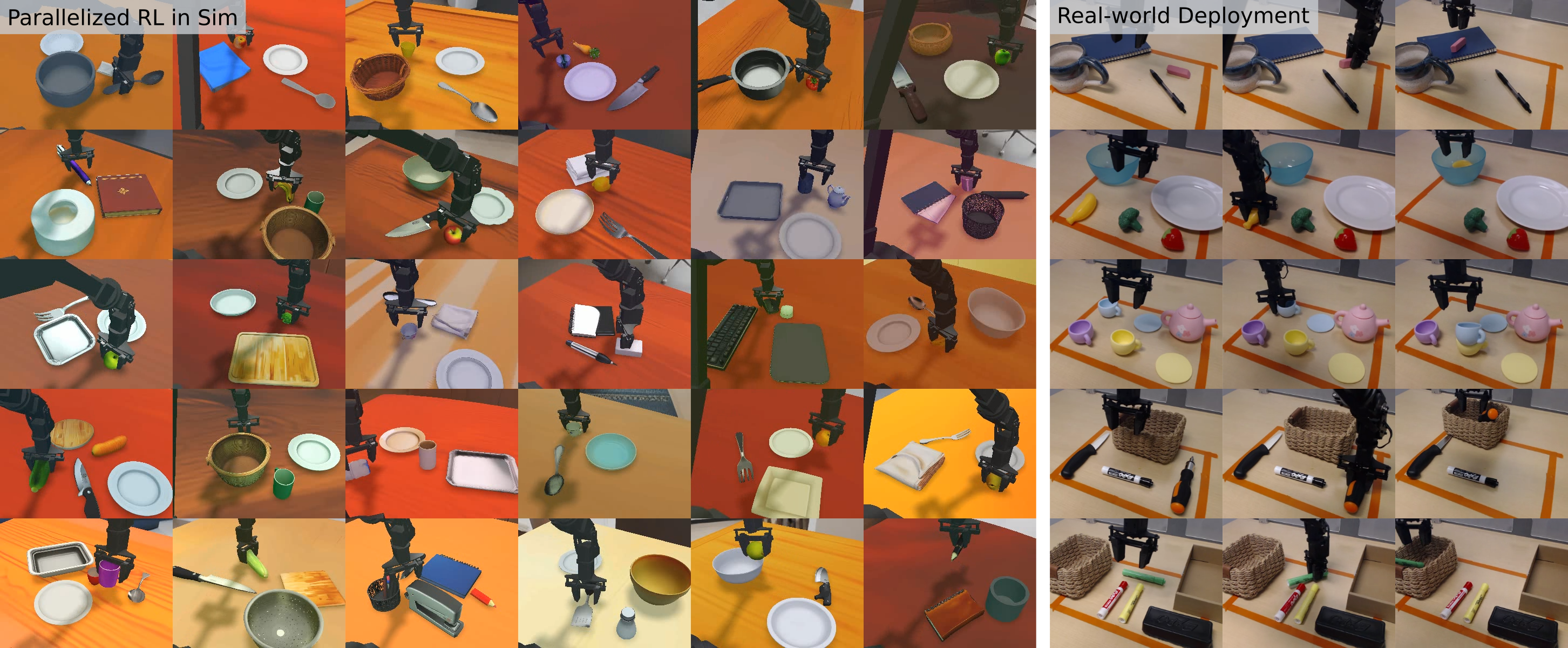}
  \caption{
    Visualizations from \cite{choi2026scaling}.
    Left: parallelized RL snapshot for training general pick-and-place using \modelname-generated scenes.
    Right: sim-to-real deployment of an \modelname fine-tuned VLA.
  }
  \label{fig:policy_learning_and_deploy}
\end{figure}

Beyond static generation quality and manual usability inspection, we examine whether generated environments support policy learning in the loop by summarizing a separate downstream study.
Choi et al.~\cite{choi2026scaling} use \modelname-generated interactive environments for online reinforcement learning (RL) of robot vision-language-action (VLA) policies, starting from a $\pi_0$-style imitation policy $\pi_{\mathrm{pre}}$ pretrained on BridgeV2~\cite{walke2023bridgedata}.
This places generated scenes inside an actual policy-optimization pipeline, where they must provide stable interaction dynamics, task diversity, and transferable learning signals rather than merely being visually plausible or physically loadable.

Table~\ref{tab:closed_loop_validation} summarizes the closed-loop results reported by the downstream study. Using only generated scenes, online RL improves simulation success from 9.7\% to 79.8\%. 
Scaling the number of generated training scenes from $N=1$ to $N=50$ increases out-of-distribution (OOD) success from 53.2\% to 77.9\% and reduces the in-distribution/out-of-distribution (ID--OOD) gap from 41.1 to 2.6 percentage points. 
In contrast, a policy trained on three hand-built SimplerEnv~\cite{li24simpler} scenes transfers poorly to \modelname scenes, achieving only 36.0\% success. 
With domain randomization, policies trained in generated environments further transfer to real robots, improving overall task success from 21.7\% to 75.0\% and reducing dynamics failures from 66.7\% to 18.3\% across 12 real-world scenes and 240 trials.

Choi and Xu~\cite{choi2026rankq} further use \modelname-generated scenes to train sim-to-real VLA policies for cube stacking via offline-to-online RL, raising real-world cube-stacking success from 43.1\% to 88.9\% across 144 trials. 
Collectively, these results position \modelname-generated environments as a scalable generative simulation substrate for closed-loop policy improvement.

\begin{table}[!htbp]
\centering
\small
\caption{\textbf{Downstream closed-loop validation of generated environments.}
Results are summarized from a large-scale sim-to-real vision-language-action (VLA) reinforcement learning study~\cite{choi2026scaling}.}
\label{tab:closed_loop_validation}
\begin{tabularx}{\linewidth}{@{}>{\raggedright\arraybackslash}p{2.75cm}
                                >{\raggedright\arraybackslash}p{4.25cm}
                                >{\raggedright\arraybackslash}X@{}}
\toprule
\textbf{Validation Axis} & \textbf{Downstream Setting} & \textbf{Key Result} \\
\midrule
Online trainability
  & Fine-tune $\pi_{\mathrm{pre}}$ using only EmbodiedGen-generated scenes
  & Simulation success improves from 9.7\% to 79.8\%, and average completion time decreases from 10\,s to 8\,s. \\
\midrule
Scene-distribution scaling
  & Scale the number of generated training scenes from $N=1$ to $N=50$
  & OOD success improves from 53.2\% to 77.9\%, and the ID--OOD gap shrinks from 41.1 to 2.6 points. \\
\midrule
Hand-built scene comparison
  & Train on three hand-built SimplerEnv scenes and evaluate on EmbodiedGen scenes
  & The policy reaches 96.7\% success on hand-built scenes, but only 36.0\% on EmbodiedGen scenes. \\
\midrule
Real-robot transfer
  & 12 real-world scenes and 240 real-robot trials
  & Real-world task success improves from 21.7\% to 75.0\%, while the dynamics failure rate decreases from 66.7\% to 18.3\%. \\
\bottomrule
\end{tabularx}
\end{table}

\FloatBarrier

\section{Related Work}
\label{sec:related_work}

Prior work has advanced individual components of generative 3D assets, scene layout, affordance annotation, natural-language editing, and embodied policy learning.
\modelname differs in treating these components as one simulation infrastructure problem: generated worlds must satisfy a sim-ready asset contract, expose interaction semantics, instantiate task-conditioned layouts, remain editable as persistent world states, and support downstream policy learning without manual scene authoring.

\paragraph{Sim-Ready 3D Asset Generation.}
3D generation has evolved rapidly from score-distillation-based optimization~\cite{poole2023dreamfusion} to feed-forward generation paradigms (Zero-1-to-3~\cite{liu2023zero123}, LRM~\cite{hong2024lrm}); state-of-the-art methods such as TRELLIS~\cite{xiang2024structured,xiang2025trellis2}, SAM3D~\cite{sam3dteam2025sam3d3dfyimages}, and Hunyuan3D~\cite{zhao2025hunyuan3d20,lai2025hunyuan3d25} now achieve end-to-end generation of high-quality textured meshes.
However, these methods optimize for visual fidelity, providing visualization-level but not simulation-level usability~\cite{hong2024lrm}.
Several works attempt to bridge this gap: Gen2Sim~\cite{katara2024gen2sim} combines diffusion-generated meshes with LLM-estimated physical parameters to support robot RL training; PhysX-3D~\cite{cao2025physx3d} augments TRELLIS with a physics VAE to produce assets with explicit mass and friction attributes; PhysX-Anything~\cite{cao2025physxanything} employs a VLM-driven pipeline to predict physical properties from a single image; PhysForge~\cite{yang2026physforge} targets interactive virtual environments and guides asset generation with physics simulation constraints.
These works move toward physical asset generation, but they do not jointly enforce the full sim-ready contract used here: quality-gated generation, mesh repair, collision proxy generation, physical metadata recovery, and standardized multi-simulator export from open-ended text or image inputs.

\paragraph{3D Indoor Scene Layout and Large-Scale Scenes Generation.}
LayoutGPT~\cite{feng2023layoutgpt} prompts language models to directly predict object bounding-box coordinates; Holodeck~\cite{yang2024holodeck} combines GPT-4 reasoning with Objaverse asset retrieval to produce navigable indoor environments; PhyScene~\cite{yang2024physcene} integrates collision, layout, and accessibility constraints within a diffusion model, representing an early effort to incorporate physical-interactivity guidance at generation time.
More recently, Rein3D~\cite{wang2026rein3d} applies reinforcement learning to refine panoramic diffusion-based indoor scene generation, while Agentic 3D Scene~\cite{liu2025agentic3d} leverages VLM agents for spatially contextualized reasoning during scene synthesis.
These methods mainly target scene plausibility or navigability, whereas \modelname starts from embodied tasks and explicitly decomposes scenes into robot, background, context, manipulated objects, and distractors before solving physical placement.
For large-scale generation, Infinigen~\cite{infinigen2024indoors} offers a constraint-driven procedural indoor generation framework, but its collision proxies are not convex-decomposed and it does not accept natural language as a control input.
EmbodiedGen V1~\cite{wang2025embodiedgengenerative3dworld} introduced panorama-based single-room background generation, but it lacks the task-level semantic decomposition, robot reachability constraints, multi-room topology, addressable furniture instances, and standardized simulator export required by the present world-generation stack.

\paragraph{3D Asset Affordance Labeling.}
Early affordance annotation relies on manual effort: 3D AffordanceNet~\cite{deng20213daff} establishes a benchmark covering 23 affordance categories, while Where2Act~\cite{mo2021where2act} learns actionable regions on articulated objects from real robot interaction trajectories---both at high annotation cost and with limited generalization to novel object categories.
Recent work scales coverage through foundation models: P3-SAM~\cite{ma2025p3sam} extends SAM to native 3D part segmentation; SegViGen~\cite{li2026segvigen} repurposes the structural knowledge embedded in 3D generative models for part segmentation; ManiTwin~\cite{wang2026manitwin} scales manipulation annotation to 100K simulation-ready assets.
The key distinction is integration: prior methods provide labels or segmentation pipelines around existing assets, while \modelname co-produces sim-ready geometry and structured part-level affordance annotations so that generated objects enter scene generation with queryable interaction semantics.

\paragraph{Natural-Language-Driven 3D Scene Editing.}
Chat-Edit-3D~\cite{fang2024chatedit3d} enables dialogue-driven iterative scene editing via visual expert models, but relies on a 2D Hash-Atlas mechanism and does not maintain a physically deployable 3D world state.
LayoutGPT~\cite{feng2023layoutgpt} and Holodeck~\cite{yang2024holodeck} regenerate the entire scene on each prompt update, precluding bounded, stateful local edits.
General-purpose coding agents~\cite{openai2025codex,google2025geminicli} lack domain-specific skills and a persistent world state, while professional authoring tools such as Blender~\cite{blender} and Maya~\cite{autodesk_maya} do not expose natural-language interfaces.
\modelname connects these two sides: language agents operate over a persistent typed world state, while domain skills commit bounded edits only after grounding, collision-aware placement, physical validation, and simulator-compatible export.

\paragraph{Embodied Policy Learning and Sim-to-Real Transfer.}
Vision-language-action (VLA) models~\cite{brohan2023rt2,kim2024openvla,black2024pi0,ye2025gigabrain,lin2026holobrain} marry pretrained vision-language backbones with robot action prediction and demonstrate strong cross-task generalization; however, online RL fine-tuning demands large quantities of physically plausible simulation environments~\cite{choi2026scaling}.
Embodied benchmarks such as RLBench~\cite{james2020rlbench} and ManiSkill3~\cite{tao2024maniskill3} provide structured evaluation tasks, but their fixed scenes and limited asset pools are ill-suited for training generalizable policies at scale.
Domain randomization~\cite{tobin2017domainrand} partially closes the visual domain gap, and works such as Embody4D~\cite{tu2026embody4d} explore richer spatiotemporal modeling through 4D world models, yet neither solves the data-supply problem of generating diverse, physically valid, task-conditioned training environments.
\modelname addresses this bottleneck from the environment side by producing sim-ready scenes that can be used directly in downstream online policy-improvement loops.

\section{Conclusion}
\label{sec:conclusion}

This work positions generative 3D world building as simulation infrastructure for embodied intelligence.
The central challenge is not only to produce plausible 3D content, but to generate worlds in which embodied agents can operate, interact, be evaluated, and learn.
Building on EmbodiedGen V1~\cite{wang2025embodiedgengenerative3dworld}, \modelname operationalizes this shift through a unified sim-ready world representation.
This representation makes metric scale, physical validity, affordances, stateful editability, and simulator interfaces persistent properties of each generated world, so that the output of one stage remains valid input for task generation, policy training, and evaluation.

\modelname realizes this view through an executable world-generation stack built on a pluggable sim-ready asset pipeline.
Rather than treating raw generative outputs as final artifacts, the asset pipeline repairs, validates, annotates, and exports them as simulator-ready entities, which then compose into language-conditioned task worlds, large-scale navigable scenes, and stateful Vibe Coding edits.
This lets users change simulation environments through controlled, physics-aware operations rather than case-by-case manual scene editing.
In this sense, the system upgrades sim-ready assets into reusable, policy-ready embodied task environments.

The evaluation shows that this design improves not only generation quality, but also downstream executability.
The asset pipeline reaches 96.5\% human acceptance and 98.6\% collision success, while 83.3\% of task-driven interactive worlds can be used for downstream simulation without manual modification.
These results indicate that the pipeline preserves both perceptual quality and execution constraints across the asset-to-world generation process.
More importantly, the generated environments contribute to policy improvement: online reinforcement learning in generated worlds raises simulation success from 9.7\% to 79.8\%, and the resulting policies improve real-robot task success from 21.7\% to 75.0\%~\cite{choi2026scaling}.
This closes a practical loop between 3D generation and embodied learning: open-ended intent can be converted into executable environments, validated through physics and interaction semantics, edited through natural language, and reused as scalable training and evaluation substrates.

Viewed more broadly, \modelname points toward a next stage of 3D generative systems measured not only by visual fidelity or diversity, but also by whether generated worlds can support closed-loop embodied behavior.
Executable world generation provides a path toward richer task curricula, broader sim-to-sim reuse, and more scalable sim-to-real policy development.
This direction can make generated 3D worlds an increasingly practical interface between human task specification, robot learning, and real-world deployment.


\newlength{\vertsep}
\setlength{\vertsep}{.085in}
\newlength{\imsize}
\setlength{\imsize}{.365\textwidth}

\clearpage

\begingroup
\hbadness=10000
\small
\setlength{\bibsep}{5pt plus 0.8ex}
\bibliography{paper}
\bibliographystyle{unsrtnat}
\endgroup

\end{document}